\documentclass[11pt]{article}

\usepackage[preprint]{acl}

\usepackage{times}
\usepackage{latexsym}

\usepackage[T1]{fontenc}

\usepackage[utf8]{inputenc}
\usepackage{booktabs,multirow}
\usepackage{microtype}

\usepackage{inconsolata}

\usepackage{graphicx}
\usepackage{makecell}
\usepackage{algorithm}
\usepackage{algpseudocode}

\usepackage{amsmath,amsfonts,bm}

\def\eqref#1{equation~\ref{#1}}

\def\1{\bm{1}}

\def\va{{\bm{a}}}

\def\vh{{\bm{h}}}

\def\vk{{\bm{k}}}

\def\vm{{\bm{m}}}

\def\vq{{\bm{q}}}

\def\vv{{\bm{v}}}

\def\vx{{\bm{x}}}
\def\vy{{\bm{y}}}
\def\vz{{\bm{z}}}

\def\mA{{\bm{A}}}
\def\mB{{\bm{B}}}
\def\mC{{\bm{C}}}

\def\mE{{\bm{E}}}

\def\mI{{\bm{I}}}

\def\mK{{\bm{K}}}
\def\mL{{\bm{L}}}
\def\mM{{\bm{M}}}

\def\mP{{\bm{P}}}
\def\mQ{{\bm{Q}}}
\def\mR{{\bm{R}}}
\def\mS{{\bm{S}}}

\def\mU{{\bm{U}}}
\def\mV{{\bm{V}}}
\def\mW{{\bm{W}}}
\def\mX{{\bm{X}}}
\def\mY{{\bm{Y}}}
\def\mZ{{\bm{Z}}}

\DeclareMathAlphabet{\mathsfit}{\encodingdefault}{\sfdefault}{m}{sl}
\SetMathAlphabet{\mathsfit}{bold}{\encodingdefault}{\sfdefault}{bx}{n}

\newcommand{\R}{\mathbb{R}}

\newcommand\ie{\textit{i.e.,}}

\newcommand{\beq}{\begin{equation}}
\newcommand{\eeq}{\end{equation}}
\newcommand{\beqnn}{\begin{equation*}}
\newcommand{\eeqnn}{\end{equation*}}
\newcommand{\beqy}{\begin{eqnarray}}
\newcommand{\eeqy}{\end{eqnarray}}
\newcommand{\beqynn}{\begin{eqnarray*}}
\newcommand{\eeqynn}{\end{eqnarray*}}
\newcommand{\bit}{\begin{itemize}}
\newcommand{\eit}{\end{itemize}}
\newcommand{\ben}{\begin{enumerate}}
\newcommand{\een}{\end{enumerate}}
\newcommand{\bex}{\begin{example}}
\newcommand{\eex}{\end{example}}

\newcommand{\balg}[1]{\begin{algorithm} \caption{#1}}
\newcommand{\ealg}{\end{algorithm}}

\newcommand{\balgc}{\begin{algorithmic}[1]}
\newcommand{\ealgc}{\end{algorithmic}}

\newcommand{\bary}{\begin{array}}
\newcommand{\eary}{\end{array}}
\newcommand{\bmx}{\begin{bmatrix}}
\newcommand{\emx}{\end{bmatrix}}
\newcommand{\bsmx}{\left[\begin{smallmatrix}}
\newcommand{\esmx}{\end{smallmatrix}\right]}
\newcommand{\bmxc}[1]{\left[\begin{array}{@{}#1@{}}}
\newcommand{\emxc}{\end{array}\right]}
\newcommand{\bcn}{\begin{center}}
\newcommand{\ecn}{\end{center}}

\usepackage{url}
\usepackage{amsmath, amssymb, amsthm, amsfonts}
\usepackage{bm}
\usepackage[capitalize,noabbrev]{cleveref}
\usepackage{wrapfig}
\usepackage{subfig}
\usepackage{inconsolata}
\usepackage{graphicx}
\usepackage{booktabs}
\usepackage{multicol, multirow}
\usepackage{enumitem}
\usepackage{amsmath,amsthm,mathtools}
\usepackage{xcolor}
\newcommand{\sol}[1]{\textcolor{blue!70!black}{#1}}   
\newcommand{\prj}[1]{\textcolor{cyan!60!black}{#1}}   
\newcommand{\spdup}[1]{\textcolor{green!40!black}{#1}}

\newcommand{\methodname}{\texttt{EvoEdit}}

\newcommand{\std}[1]{\,{\scriptsize$\pm$\,#1}}

\theoremstyle{plain}
\newtheorem{theorem}{Theorem}[section]
\newtheorem{proposition}[theorem]{Proposition}
\newtheorem{lemma}[theorem]{Lemma}
\newtheorem{corollary}[theorem]{Corollary}
\theoremstyle{definition}

\theoremstyle{remark}
\newtheorem{remark}[theorem]{Remark}

\makeatletter
\let\oldtheequation\theequation
\renewcommand\tagform@[1]{\maketag@@@{\ignorespaces#1\unskip\@@italiccorr}}
\renewcommand\theequation{(\oldtheequation)}
\makeatother
\creflabelformat{equation}{#2\textup{#1}#3}  

\title{EvoEdit: Evolving Null-space Alignment for Robust and Efficient Knowledge Editing}

\author{
  \textbf{Sicheng Lyu\textsuperscript{\rm 1,2,3}}\thanks{Equal contribution.} \,
  \textbf{Yu Gu\textsuperscript{\rm 1}}\footnotemark[1] \,
  \textbf{Xinyu Wang\textsuperscript{\rm 1,3}} \,
  \textbf{Jerry Huang\textsuperscript{\rm 2,4}} \\
  \textbf{Sitao Luan\textsuperscript{\rm 2,4}} \,
  \textbf{Yufei Cui\textsuperscript{\rm 1,2}} \,
  \textbf{Xiao-Wen Chang\textsuperscript{\rm 1}} \,
  \textbf{Peng Lu\textsuperscript{\rm 4}}\thanks{Corresponding author.}
  \\
  \textsuperscript{\rm 1}McGill University \quad
  \textsuperscript{\rm 2}Mila - Quebec AI Institute \quad
  \textsuperscript{\rm 3}SimpleWay.AI \quad
  \textsuperscript{\rm 4}Universit\'e de Montr\'eal \\
  \texttt{sicheng.lyu}@\texttt{mail.mcgill.ca} \quad \texttt{yu.gu4@mail.mcgill.ca},\quad
  \texttt{peng.lu@umontreal.ca}
}

\begin{document}

\maketitle

\begin{abstract}
Ensuring that Large Language Models (LLMs) remain useful throughout their lifetimes requires frequent updates to maintain factual accuracy, yet standard ``locate-then-edit'' methods suffer from catastrophic interference in sequential settings, where successive updates degrade previously integrated knowledge. In this work, we propose {\methodname}, a framework that achieves stable, large-scale sequential editing through \emph{sequential null-space alignment}. Unlike previous works, \methodname{} dynamically projects knowledge edits into the null space of both original and previously modified knowledge, with theoretical guarantees of output invariance across extended edit sequences. This approach balances model adaptability with knowledge preservation, preventing the cumulative interference that commonly leads to model collapse with previous methods. Empirical evaluations across multiple LLMs demonstrates state-of-the-art performance at moderate scales (e.g., 2\textsf{K} edits) and provides substantial improvements in stability and retention at large scales of up to 10\textsf{K} edits. Furthermore, the method achieves up to a \emph{3.5$\times$ speedup} over existing baselines, establishing \methodname{} as a principled, efficient, and scalable solution for reliable knowledge management in real-world LLM deployments. Code is available at \href{https://github.com/Euphoria040201/EvoEdit}{here}.
\end{abstract}
\section{Introduction}

Large language models (LLMs) have demonstrated a remarkable ability to store and recall vast amounts of knowledge during pre-training~\citep{gpt-4, claude3, llama3, qwen-2.5,  deepseek-r1, gemini-2.5, qwen-3}, enabling them to utilize this knowledge for downstream tasks. 
However, powerful as they may be, becoming outdated remains a concern, with their static knowledge leading to eventual hallucinations or factual errors over time~\citep{de-cao-etal-2021-editing, MEND}. While re-training with updated data offers a straightforward solution, the computational expensive and risks of overfitting or catastrophic forgetting~\citep{forgetting} prohibits this as an effective approach. Alternatively, \emph{model editing} has emerged as a targeted updating procedure of specific factoids without additional training~\citep{wang2024knowledge, gupta-etal-2024-model}. 

Most editing methods follow a \emph{locate-then-edit paradigm}~\citep{ROME, MEMIT}, which first identifies a small set of influential parameters within the model and then applies a perturbation to integrate new knowledge. 
While effective at isolated edits, recent studies~\citep{mirage, MEMOIR} reveal that these approaches suffer from catastrophic interference when deployed in sequential editing scenarios, where multiple updates are applied over time~\citep{yang-etal-2024-butterfly}. 
As new perturbations accumulate, previous edits are also disrupted, leading to greater difficulty in preserving knowledge and more severe issues such as model collapse. Thus a critical barrier remains when deploying model editing in realistic settings; continual updates are essential for keeping LLMs reliable and up-to-date. Many methods fail to capture this real-world complexity and are only evaluated in synthetic frameworks~\citep{yao-etal-2023-editing, wang2024knowledge}. Thus under more realistic conditions which better mimic the ever-changing world~\citep{wang2024knowledge}, performance of existing methods crater, revealing that current approaches {degrade catastrophically after only a few hundred edits} and fail to scale to the real-world.

This paper addresses the challenge of \emph{scalable sequential factual knowledge editing} in large language models. We propose \methodname, a sequential null-space alignment framework designed for high-volume updates to a model’s stored facts. Empirical evaluations across multiple benchmarks show that \methodname{} matches state-of-the-art editing performance while significantly improving knowledge retention and achieving up to a \emph{3.5$\times$ speedup}. 
We summarize our key contributions as follows:
\begin{itemize}
\item We propose a robust and efficient subspace alignment method that mitigates null-space drift caused by sequential knowledge updates.
\item We provide theoretical guarantees of output invariance across extended edit sequences through a provable null-space projection mechanism.
\item We develop a numerically stable and computationally efficient implementation leveraging the Woodbury matrix identity for scalable updates.
\end{itemize}

\section{Related Work}

\subsection{Evaluation of Model Editing}
Current evaluations of model editing primarily focus on assessing both the effectiveness of edits and their adverse impacts on overall model performance. Effectiveness is typically measured along three key dimensions:  
1) \textit{Reliability} — the success rate of applying the intended knowledge edits;  
2) \textit{Generalization} — the ability of the edited knowledge to remain consistent under paraphrased or semantically similar inputs; 3) \textit{Locality} — the extent to which the edit influences unrelated/peripheral knowledge. These metrics collectively capture whether an edit operates within its intended scope~\citep{ROME, easyedit, yao-etal-2023-editing}. Beyond these, recent work has also examined the fine-grained effects of post-editing, such as mitigating biases or preventing the injection of harmful information into large language models (LLMs)~\citep{chen2024large, chen2024editingllmsinjectharm}.

\subsection{Taxonomy of Model Editing }
The current model editing approaches can be categorized as either introducing an auxiliary mechanism or updating (a subset of) the model’s parameters to store new knowledge~\citep{wang2024knowledge}. This paper focuses on the latter and follows the locate–then–edit paradigm proposed by \texttt{ROME}~\citep{ROME}. \texttt{MEMIT}~\citep{MEMIT} extends this to support batched edits. Other methods adopt meta-learning, exemplified by MEND~\citep{MEND} and KE~\citep{de-cao-etal-2021-editing}, which learn to predict parameter updates.

\section{Preliminaries}

Model editing~\citep{ROME, MEMIT, Alphaedit} seeks to update factual knowledge encoded in a frozen language model through one-shot or batched parameter updates.  
Facts are commonly represented as triples \((s,r,o)\), where \(s\) denotes a subject, \(r\) a relation, and \(o\) an object. For example, an arbitrary fact can be stored as
\[
(s,r,o) = \left(\text{``CEO of Tesla''},\; \text{``is''},\; \text{``Tom Zhu''}\right).
\]
Given a prompt containing the pair \((s,r)\), the model is expected to generate the token(s) corresponding to \(o\).  
If this fact is edited, the same prompt should instead produce a new target object \(\tilde{o}\), e.g.,
\[
(s,r,\tilde{o}) = \left(\text{``CEO of Tesla''},\; \text{``is''},\; \text{``Elon Musk''}\right).
\]
Ideally, the change induced from $o$ to $\tilde{o}$ for a specific fact should not impact un-related or ancillary facts, \ie{} when updating $(s_1, r_1, o_1)$ to $(s_1, r_1, \tilde{o}_1)$ $(s_2, r_2, o_2)$ should not be influenced if the relationship between these two facts is weak. 

\subsection{Feed-Forward Blocks as Associative Memory}

We consider auto-regressive LLMs, where the next token \(x_t\) is predicted from its preceding context.  
Let \(\vh^{(l-1)}\) denote the hidden state input to layer \(l\) of \(x_t\). The feed-forward network (FFN) computes
\[
\underbrace{\vm^{\left(l\right)}}_{\vv} \;=\; \mW^{\left(l\right)}_{\mathrm{out}} \;
   \underbrace{\sigma\!\left(\mW^{\left(l\right)}_{\mathrm{in}}\,
   (\vh^{(l-1)} + \va^{\left(l\right)})\right)}_{\vk},
\]
where \(\va^{\left(l\right)}\) is the output of the attention block, \(\mW^{\left(l\right)}_{\mathrm{in}}\) and \(\mW^{\left(l\right)}_{\mathrm{out}}\) are the learnable projection matrices, and \(\sigma\) is the activation function.  
This transformation can be interpreted as an associative memory~\citep{FFN-KV-memory, softmax-linear-units}.  
The nonlinear projection \(\sigma\left(\mW^{\left(l\right)}_{\mathrm{in}}\!\left(\vh^{\left(l-1\right)}+\va^{\left(l\right)}\right)\right)\) generates a key representation $\vk$ that encodes the input context, while the output projection \(\mW^{\left(l\right)}_{\mathrm{out}}\) maps this key to a corresponding value $\vv$. When editing knowledge, each update can be represented as a key–value pair, where $\vk$ encodes \((s,r)\) and $\vv$ encodes the new target object \(\tilde{o}\). This associative interpretation provides a principled rationale for knowledge editing: modifying either the key space or the output projection directly alters how facts are retrieved and expressed.  
In locate-then-edit frameworks, for instance, the goal is to update $\mW_{\mathrm{out}}^l$ to reflect the revised associations. 
For notational simplicity, we omit layer-specific subscripts and superscripts of $\mW_{\mathrm{out}}^l$ in what follows.


\subsection{Model Editing in LLMs}
In the locate-then-edit paradigm, model parameters $\mW$ are perturbed by a small update $\bm{\Delta}$ to reflect the knowledge update. Specifically, given a knowledge triple $(s, r, \tilde{o})$, the modified weights $\mW + \bm{\Delta}$ are expected to encode the corresponding new key–value association, ensuring $(\mW + \bm{\Delta})\vk = \vv$ ideally. The central challenge lies in determining an optimal $\bm{\Delta}$: one that effectively injects or replaces the target knowledge while minimizing unintended interference with knowledge that should be preserved~\citep{MEMIT}. This can be formulated as the following optimization problem.
\begin{equation}
    \resizebox{0.85\linewidth}{!}{$\underset{{\bm{\Delta}}}{\min}
(
\underbrace{\left\|\left(\mW+{\bm{\Delta}}\right)\mK_{1} - \mV_{1} \right\|^{2}}_{{\color{blue}(1)}\,\textrm{Ensure effective edits}} + \underbrace{\left\| \left(\mW+{\bm{\Delta}}\right)\mK_{0} - \mV_{0} \right\|^{2}}_{{\color{blue}(2)}\,\textrm{Preserve model knowledge}}),$}
\end{equation}
where $\mW\!\in\!\R^{d_v \times d_K}$, $\mK_1\!\in\!\R^{d_K \times u}$ and $\mV_1\!\in\!\R^{d_v \times u}$ denote the matrices by collecting the $\vk$ and $\vv$ of renewing knowledge, $\mK_0\!\in\!\R^{d_K \times N}$ and $\mV_0\!\in\! \R^{d_v \times N}$ denote the matrices by collecting the $\vk$ and $\vv$ of the knowledge we would like to preserve. $\mK_0$ is usually estimated by utilizing sufficient text input~\citep{MEMIT}, e.g., 100K $(s, r, o)$ triplets from Wikipedia are chosen randomly to encode $\mK_0$.Joint optimization reveals a fundamental preservation-injection dilemma: parameter updates for new facts often disrupt retained knowledge. To resolve this, recent null-space projection methods have emerged to mitigate such interference.

\subsection{Null-space of Preserved Knowledge}
\paragraph{Null-space Projection.} Given two matrices $\mA$ and $\mB$, if 
$\mB^\top\mA = \bm{0}$, then by definition the columns of $\mB$ lie in the left null space of $\mA$ (or equivalently the (right) null-space of $\mA^\top$). 
In the context of model editing, consider an update $\bm{\Delta}$ projected onto the left null space of $\mK_0$, i.e., $\bm{\Delta}\mP \mK_0 = \bm{0}$.
Under this constraint, we obtain
\begin{align}
(\mW + \bm{\Delta}\!\mP)\mK_0 = \mW\!\mK_0 = \mV_0.
\end{align} 
This ensures that the update $\bm{\Delta}\mP$ leaves the existing key–value mappings $\mK_0 \mapsto \mV_0$ unchanged.
Thus, if the null-space projector $\mP$ of the historical knowledge matrix $\mK_0$ is available, any update can be projected through $\mP$ to ensure that $\bm{\Delta}\mP$ preserves the existing mappings, thereby safeguarding preserved knowledge from interference.

\paragraph{Null-space Projector Estimation.}
Directly computing the null-space projector $\mP$ of the historical knowledge key matrix $\mK_0$ is intractable, since $\mK_0$ typically possesses a growing number of columns as knowledge editing progresses and its column vectors are rarely orthonormal to each other. To address this, we exploit the fact that $\mK_0$ and its non-central covariance matrix $\mK_0\mK_0^\top$ share a left null space,  i.e.,Null$(\mK_0^T)$ = Null$(\mK_0\mK_0^T)$. Thus the null space can be estimated via $\mK_0\mK_0^\top$, avoiding materializing the dense matrix $\mK_0$ and reducing memory cost because the row dimension is fixed and typically much smaller than the column dimension~\citep{null-space-corvariance, Alphaedit, multilingual-alpha-edit}. One way of obtaining the (approximate) left null-space basis is first to perform singular value decomposition (SVD) on the covariance matrix:
\begin{equation}
\begin{split}
& \{\mU, \bm{\Lambda}, \mU^\top\}= \mathrm{SVD}\!\left(\mK_0\mK_0^\top\right), \\
& \bar{\mU} = \mU_{[:,i:d_K]}, 
\ \ \bm{\Lambda}_{[i]} < \tau \leq \bm{\Lambda}_{[i-1]},
\end{split}
\end{equation}
where $\tau$ is a threshold  (e.g., 10\textsuperscript{-2}), and 
$\bar{\mU}$ is formed by the singular vectors associated with singular values smaller than $\tau$.
The threshold is necessary because singular values are rarely exactly zero in practice. 
The resulting (approximate) null-space orthogonal projector is then defined as
\begin{align}
\mP = \bar{\mU}\bar{\mU}^\top.
\end{align}
\begin{figure*}[t!]
    \centering
    \includegraphics[width=0.95\linewidth]{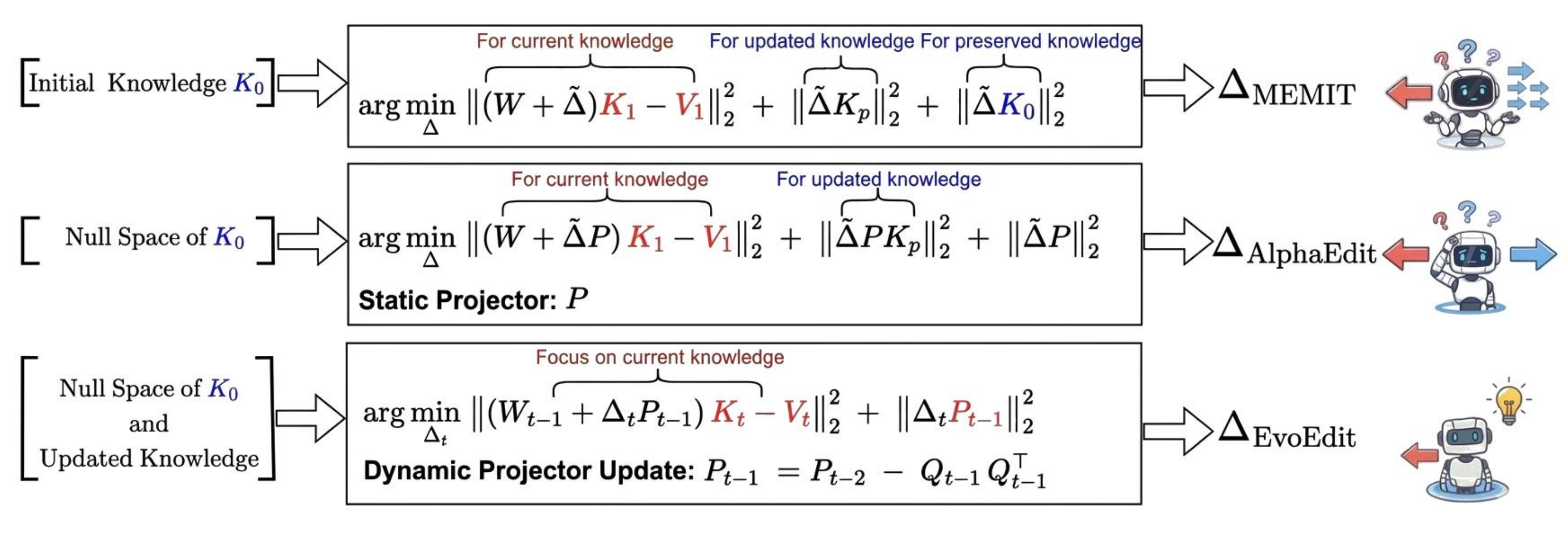}
    \caption{Evolution of Model Editing Objectives. Comparison between \texttt{MEMIT} (top), \texttt{AlphaEdit} with a static projector (middle), and the proposed \methodname{} (bottom), which utilizes a Dynamic Projector Update to sequentially refine knowledge while preserving the null space of previous states.}
    \label{fig: evoeit}
\end{figure*}
\section{Methodology}
\label{sec:method}
We begin by examining how sequential knowledge updates induce null-space drift, a key challenge that undermines the stability of continual editing. To address this, we propose \methodname—a novel framework that incrementally aligns the null space through a stable, synergistic mechanism, ensuring that each new edit integrates cleanly without degrading prior updates.
\subsection{Null-Space Drift and Interference}

During sequential editing, the goal is to determine a series of perturbations 
\(\{\bm{\Delta}_1, \dots, \bm{\Delta}_t\}\) corresponding to knowledge updates 
\(\{\mK_1, \dots, \mK_t\}\). Two challenges exist: not only must the initial preserved knowledge \(\mK_0\) remain intact, but the newly injected knowledge should also remain robust against contamination from subsequent updates.  Prior work, including \texttt{AlphaEdit}~\citep{Alphaedit} and \texttt{LangEdit}~\citep{multilingual-alpha-edit}, demonstrates the benefits of performing updates within the null space of preserved knowledge. \texttt{AlphaEdit}, however, employs a fixed null-space projector, ignoring the drift induced by sequential updates. \texttt{LangEdit} mitigates this by recomputing the null space after each edit, but its reliance on the SVD of the non-centered covariance matrix \(\widehat{\mK}_t \widehat{\mK_t}^\top\), where \(\widehat{\mK}_t = [\mK_0, \dots, \mK_t]\), however, because $\widehat{\mK}$ is rank-deficient, Performing Singular Value Decomposition (SVD) on this ill-conditioned matrix yields highly unstable results, as the disproportionately wide range of magnitudes causes small singular values to be lost to floating-point roundoff errors, while the corresponding singular vectors become significantly distorted and cease to accurately span the intended subspace.

\subsection{Efficient Null-space Alignment}
We now describe sequential editing as an optimization problem. Given the model weights $\mW_{t-1}$ and the collection of all initial and updated key-value pairs $\left(\widehat{K}_t, \widehat{V}_t\right)$, our goal is to determine the optimal perturbation $\hat{\Delta}$ to inject $\left(\mK_t, \mV_t\right)$ by solving the following problem at time step $t$:
\begin{equation}
  \label{eq:obj_seq_1}
\begin{split}
&    \underset{\bm{\Delta}_{t}}{\min}\left\|\left(\mW_{t-1} + \hat{\bm{\Delta}}\right)\mK_t\!-\!\mV_t\right\|^2 \\
& + \left\|\left(\mW_{t-1} + \hat{\bm{\Delta}}\right)\!\widehat{\mK}_{t-1}-\widehat{\mV}_{t-1}\right\|^2,
\end{split}  
\end{equation}
where $\widehat{\mV}_{t-1} = [\mV_0, \dots, \mV_{t-1}]$ denotes the collection of all initial and updated values up to step $t-1$. To address this, we require the null-space projector of $\widehat{\mK}_{t-1}$. Instead of recomputing it from $\widehat{\mK}_{t-1}\widehat{\mK}_{t-1}^\top$, we introduce a sequential alignment approach that updates the null-space projection based on the new key matrix $\mK_{t-1}$.
\begin{figure}[th!]
    \centering
\includegraphics[width=0.9\linewidth]{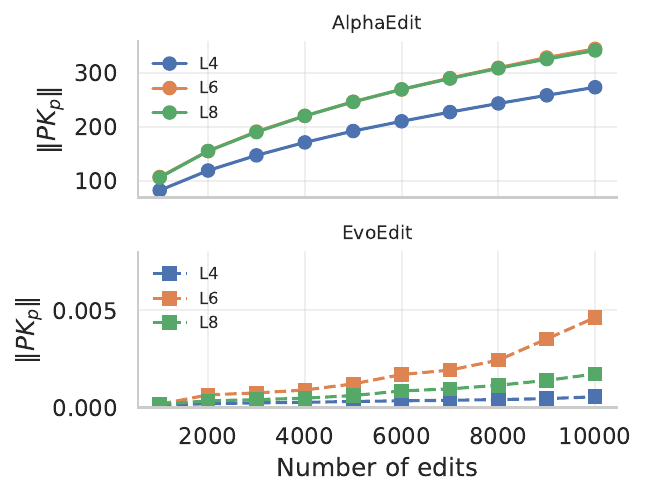}
    \caption{{Quantitative Analysis of Null-space Drift.} The growth of $\|PK_p\|_F$ in \texttt{AlphaEdit} (solid) across different layers indicates significant drift as the number of edits increases, which forces a trade-off between new knowledge acquisition and interference suppression. \methodname{} (dashed) eliminates this drift by dynamically adapting the projector, maintaining stable geometry even under extended sequential editing (up to 10k edits).}
    \label{fig:null-space-drift}
\end{figure}

Denote by $\mP_{t-2}\in\R^{d_K\times d_K}$ the orthogonal projector onto the left null space of $\widehat{\mK}_{t-2}$. To capture the portion of this null space that aligns with the basis directions induced by $\mK_{t-1}$, we perform a singular value decomposition (SVD) w.r.t. $\mP_{t-2}\mK_{t-1}$, and select singular vectors with singular values greater than a pre-defined threshold $\tau$:
\begin{equation}
\begin{split}
\{\mU, \bm{\Sigma}, \mU^\top\} = \mathrm{SVD}\!\left(\mP_{t-2}\mK_{t-1}\right), \\
\mQ_{t-1} = \mU_{[:, :i]}, \quad \Sigma_{[i]} > \tau > \Sigma_{[i-1]}.
\label{eq:svd}
\end{split}
\end{equation}
Because $\mK_{t-1}$ has far fewer columns than $\mK_0$ or $\widehat{\mK}_{t-2}$, this SVD step is both more efficient and numerically stable than recomputing the full covariance matrix. 
The updated projector is then obtained via the deflation step
\begin{equation}
\;\mP_{t-1}\;=\;\mP_{t-2}\;-\;\mQ_{t-1}\mQ_{t-1}^{\top}\;.
\label{eq:deflate_update}
\end{equation}
where $\mP_0$ is the null-space projector w.r.t. the initial preserved knowledge $\mK_0$. $\mQ_{t-1}$ indicates the directions related to the newly updated knowledge $\mK_{t-1}$. $\mP_{t-1}$ is the aligned projector  that satisfies $\mP_{t-1}\widehat{\mK}_{t-1} = \bm{0}$.

\subsection{Sequential Editing via aligned Null-space Projector}
Next, we present how to leverage the aligned projector to solve the sequential editing problem. We input $\hat{\bm{\Delta}}$ with $\bm{\Delta}\mP_{t-1}$ to \cref{eq:obj_seq_1}:
\begin{equation}
\label{eq:obj_seq}
\begin{split}
&    \underset{\bm{\Delta}_{t}}{\min}\!\left\|\left(\mW_{t-1}\!+\!\bm{\Delta}_t\mP_{t-1}\right)\mK_t\!-\!\mV_t\right\|^2 \\
&+ \left\|\left(\mW_{t-1}\!+\!\bm{\Delta}_t\mP_{t-1}\right)\!\widehat{\mK}_{t-1}\!-\!\widehat{\mV}_{t-1}\right\|^2,
\end{split}
\end{equation}
Since the projector $\mP_{t-1}$ guarantees that $\bm{\Delta}_t\!\mP_{t-1}\widehat{\mK}_{t-1} = \bm{0}$, the second term in \cref{eq:obj_seq} becomes independent of $\bm{\Delta}_t$ and can therefore be omitted. To further stabilize convergence, we introduce a regularization term $\|\bm{\Delta}_t\!\mP_{t-1}\|^2$, yielding the final optimization problem:
\begin{equation}
\resizebox{0.85\linewidth}{!}{
    $\underset{\bm{\Delta}_{t}}{\min}\!
        \left\|\left(\mW_{t-1}+\bm{\Delta}_{t}\mP_{t-1}\right)\mK_{t}-\mV_{t}\right\|^{2}
        +\left\|\bm{\Delta}_{t}\mP_{t-1}\right\|^{2}$
}.
\end{equation}

Defining the residual of the current edit as $\mR_t = \mV_t\!-\!\mW_{t-1}\mK_t$ for notational simplicity, we solve the optimization problem using the normal equations~\citep{lang2012introduction, strang2022introduction}, yielding
\begin{equation}
\label{eq:solution1}
\resizebox{0.85\linewidth}{!}{
    $\bm{\Delta}_{t}\mP_{t-1}
    = \mR_t\!\mK_{t}^{\top}\!\mP_{t-1}\left(\mK_{t}\mK_{t}^{\top} \mP_{t-1} + \mI \right)^{-1}$
}.
\end{equation}

Direct computation of \cref{eq:solution1} requires $O\!\left(d_K^3\right)$ flops due to the matrix inversion; thus, we leverage the Woodbury matrix identity~\citep{guttman1946enlargement, woodbury1950inverting, hager1989updating,  higham2002accuracy} for a more efficient formulation. Full derivations are provided in~\cref{sec: solution-proofs}.
\begin{equation}
\label{eq:solution2}
\resizebox{0.85\linewidth}{!}{   
    $\bm{\Delta}_{\texttt{EvoEdit}}= \mR_t\left(\!\mK_t^\top \mP_{t-1} \mK_t \!+\mI_r \right)^{-1}\mK_t^\top \mP_{t-1}$
}.
\end{equation}
\cref{alg:delta_computation} provides a numerically stable computation method for~\cref{eq:solution2}. We provide the complete algorithm of \methodname{} in \cref{alg:evoedit}.
\paragraph{Complexity analysis.} The per-edit complexity of \methodname{} is $O\!(d_K(rn + n^2) + n^3)$, where $n\!=\!\max(n_t, n_{t-1})$. By representing the orthogonal projector in low-rank form as $\mP\!=\!I_{d_K}\!-\!\mQ\mQ^\top$ with $\mQ\!\in\!\mathbb{R}^{d_K \times\!r}$, the method eliminates the $O\!(d_K^3)$ bottleneck typical of standard null-space approaches. Through SVD-based alignment and Woodbury-style rearrangements, the hidden dimension $d_K$ appears only linearly in the total cost. Since the cubic dependency is confined strictly to the edit size $n_t$---which is typically several orders of magnitude smaller than $d_K$---\methodname{} enables highly efficient updates without the need to materialize or invert dense $d_K\!\times\!d_K$ matrices.

\paragraph{Null-space drift comparison.} In \texttt{AlphaEdit}, the regularization term $\left\| P K_p \right\|_F$ grows with the number of edits, increasingly constraining new updates and forcing a trade-off between fitting $(W\!+\!\tilde{\Delta}P)K_1 \!\approx\!V_1$ and suppressing interference. \methodname{} avoids this bottleneck by dynamically adapting the projector. As shown in Figure~\ref{fig:null-space-drift}, the Frobenius norm of $P K_p$ in \texttt{AlphaEdit} becomes several orders of magnitude larger than in \methodname{} across layers, indicating significant null-space drift, whereas \methodname{} maintains stable geometry under extended edits.
\subsection{Theoretical Analysis}
We analyze the \emph{projector-alignment mechanism} and show that the iterate $\mP_{t-1}$ serves as (or closely approximates) the orthogonal projector onto the null space of the accumulated knowledge matrix $\widehat{\mK}_{t-1} = [\,\mK_0, \mK_1, \ldots, \mK_{t-1}\,]\!\in\!\R^{d \times r_{t-1}}$. In the exact (non-truncated) setting, the null space of $\mP_{t-1}$ coincides with the column space of $\widehat{\mK}_{t-1}$:
\[
\mathrm{Null}\!\left(\mP_{t-1}\right) = \mathrm{Range} (\widehat{\mK}_{t-1}),
\]
implying that $\mP_{t-1}$ projects onto the subspace orthogonal to all previously edited knowledge. Under the practical truncation scheme motivated by numerical stability, this equivalence holds approximately, and the deviation is provably bounded (see \cref{thm:global_error}). All proofs are provided in \cref{sec:proofs}. First, define the step-$j$ projected keys
\[
\mR_j \coloneq \mP_{j-1}\mK_j.
\]
\begin{table*}[th!]
\centering
\large
\renewcommand{\arraystretch}{1.2}
\resizebox{0.95\linewidth}{!}{
\begin{tabular}{cc|ccccc|ccc}
\toprule[1.5pt]
\multirow{2}{*}{\textbf{Method}} & \multirow{2}{*}{{\textbf{Model}}}  & \multicolumn{5}{c|}{\textsc{\textbf{Counterfact}}} & \multicolumn{3}{c}{\textsc{\textbf{ZsRE}}} \\
\cmidrule(lr){3-7} \cmidrule(lr){8-10}
&& \textbf{Eff.$\uparrow$} & \textbf{Gen.$\uparrow$} & \textbf{Spe.$\uparrow$} & \textbf{Flu.$\uparrow$} & \textbf{Consis.$\uparrow$} & \textbf{Eff.$\uparrow$} & \textbf{Gen.$\uparrow$} & \textbf{Spe.$\uparrow$} \\
\midrule[1pt]

\texttt{FT} & \multirow{9}{*}{\rotatebox{90}{\texttt{Llama-3-8B}}} & {83.33\std{0.37}} & {67.79\std{0.40}} & {46.63\std{0.37}} & {233.72\std{0.22}} & {8.77\std{0.05}} & {30.48\std{0.26}} & {30.22\std{0.32}} & {15.49\std{0.17}}\\
\texttt{MEND} & & {63.24\std{0.31}} & {61.17\std{0.36}} & {45.37\std{0.38}} & {372.16\std{0.80}} & {4.21\std{0.05}} & {0.91\std{0.05}} & {1.09\std{0.05}} & {0.53\std{0.02}}\\
\texttt{InstructEdit} & & {66.58\std{0.24}} & {64.18\std{0.35}} & {47.14\std{0.37}} & {443.85\std{0.78}} & {7.28\std{0.04}} & {1.58\std{0.04}} & {1.36\std{0.08}} & {1.01\std{0.05}}\\
\texttt{ROME} & & {64.40\std{0.41}} & {61.42\std{0.42}} & {49.44\std{0.38}} & {449.06\std{0.26}} & {3.31\std{0.02}} & {2.01\std{0.07}} & {1.80\std{0.07}} & {0.69\std{0.03}}\\
\texttt{MEMIT} & & {65.65\std{0.47}} & {64.65\std{0.42}} & {51.56\std{0.38}} & {437.43\std{1.67}} & {6.58\std{0.11}} & {34.62\std{0.36}} & {31.28\std{0.34}} & {18.49\std{0.19}} \\
\texttt{PRUNE} & & {68.25\std{0.46}} & {64.75\std{0.41}} & {49.82\std{0.36}} & {418.03\std{1.52}} & {5.90\std{0.10}} & {24.77\std{0.27}} & {23.87\std{0.27}} & {20.69\std{0.23}} \\
\texttt{RECT} & & {66.05\std{0.47}} & {63.62\std{0.43}} & {61.41\std{0.37}} & {526.62\std{0.44}} & {20.54\std{0.09}} & {86.05\std{0.23}} & {80.54\std{0.27}} & {31.67\std{0.22}} \\
\texttt{AlphaEdit} & & \underline{98.90\std{0.10}} & \underline{94.22\std{0.19}} & \underline{67.88\std{0.29}} & \underline{622.49\std{0.16}} & \underline{32.4\std{0.11}} & \underline{94.47\std{0.13}} & \underline{91.13\std{0.19}} & \textbf{32.55\std{0.22}}\\
\methodname~(Ours) & & \textbf{99.67\std{0.08}} & \textbf{94.93\std{0.27}} & \textbf{69.99\std{0.59}} & \textbf{623.09\std{0.98}} & \textbf{32.64\std{0.34}} & \textbf{95.74\std{0.03}} & \textbf{92.13\std{0.23}} & \underline{32.41\std{0.30}}\\
\midrule[1pt]

\texttt{ROME} & \multirow{4}{*}{\rotatebox{90}{\texttt{Qwen-7B}}} & {68.65\std{1.09}} & {62.44\std{0.86}} & {51.35\std{0.94}} & {539.77\std{26.09}} & {4.26\std{1.20}} & {18.93\std{2.46}} & {17.20\std{3.06}} & {7.56\std{1.63}}\\
\texttt{MEMIT} & & {65.65\std{0.47}} & {64.65\std{0.42}} & {51.56\std{0.38}} & {437.43\std{1.67}} & {6.58\std{0.11}} & {34.62\std{0.36}} & {31.28\std{0.34}} & {18.49\std{0.19}} \\
\texttt{AlphaEdit} & & \textbf{99.57\std{0.10}} & \underline{79.81\std{1.05}} & \underline{82.65\std{0.13}} & \underline{626.80\std{0.33}} & \underline{30.98\std{0.19}} & \textbf{99.74\std{0.14}} & \underline{91.58\std{0.44}} & \underline{41.58\std{0.34}}\\
\methodname~(Ours) & & 99.50\std{0.20} & \textbf{82.50\std{1.55}} & \textbf{82.73\std{0.08}} & \underline{626.80\std{0.23}} & \textbf{31.22\std{0.09}} & \underline{99.54\std{0.17}} & \textbf{91.67\std{0.72}} & \textbf{41.97\std{1.17}}\\
\midrule[1pt]

\texttt{ROME} & \multirow{4}{*}{\rotatebox{90}{\texttt{Llama-3-3B}}} & {69.88\std{2.07}} & {63.79\std{1.67}} & {48.88\std{0.65}} & \textbf{{656.96\std{3.04}}} & {1.65\std{0.75}} & {2.25\std{0.80}} & {2.03\std{0.70}} & {0.05\std{0.08}}\\
\texttt{MEMIT} & & {76.03\std{0.86}} & {77.40\std{1.12}} & {60.70\std{0.32}} & {576.23\std{8.92}} & {20.76\std{0.37}} & {0.00\std{0.00}} & {0.00\std{0.00}} & {1.59\std{1.56}}\\
\texttt{AlphaEdit} & & \underline{99.42\std{0.19}} & \underline{96.69\std{0.19}} & \underline{64.71\std{0.26}} & 624.84\std{6.66} & \underline{32.92\std{1.17}} & \underline{94.44\std{0.10}} & \underline{89.87\std{0.31}} & \textbf{30.09\std{0.13}}\\
\methodname~(Ours) & & \textbf{99.77\std{0.08}} & \textbf{97.27\std{0.08}} & \textbf{71.32\std{0.50}} & \underline{630.71\std{2.28}} & \textbf{34.10\std{0.64}} & \textbf{94.97\std{0.10}} & \textbf{89.92\std{0.49}} & \underline{29.26\std{0.22}}\\
\bottomrule[1.5pt]
\end{tabular}
}
\caption{\footnotesize Comparison of \texttt{EvoEdit} with existing methods on the sequential model editing task. We evaluate across five dimensions: \textit{Eff.} (Efficacy), \textit{Gen.} (Generalization), \textit{Spe.} (Specificity), \textit{Flu.} (Fluency), and \textit{Consis.} (Consistency), capturing both the accuracy of edits and their quality in natural language generation. The best-performing results are highlighted in bold, while the second-best results are underlined, facilitating a clear visual comparison of relative performance across methods.}
\label{tab:overall_comp}
\end{table*}
\begin{theorem}[Exact equivalence without truncation]
\label{thm:equivalence}
Suppose that at each step $j$, the update uses $\mQ_j$ as any orthonormal basis of
$\mathrm{Range}\!\left(\mR_j\right)=\mathrm{Range}\!\left(\mP_{j-1}\mK_j\right)$, i.e., \emph{no truncation is applied and all nonzero
left singular directions of $\mR_j$ are retained}. Then, for all $t\ge 0$,
\[
\mathrm{Null}\!\left(\mP_{t-1}\right)=\mathrm{Range}\left(\widehat{\mK}_{t-1}\right).
\]
\end{theorem}
For more general cases, we also provide the theoretical bound of the deviation between the truncated projector and the ideal projector.
\begin{theorem}[Global error bound with truncation]
\label{thm:global_error}
For any $t\ge 1$, let $\widetilde \mP_t$ be the projector obtained by applying truncation 
thresholds $\{\tau_j\}_{j=1}^t$. Then the cumulative deviation satisfies
\begin{align*}
&\left\|\widetilde \mP_t\!-\!\mP_t^\star\right\|_2
\;\le \\ 
&\min\!\left\{\,1,\;
\sum_{j=1}^t \frac{\left\|\mE_j\right\|_2}{\sigma_{q_j}-\sigma_{q_j+1}} \;+\;
\max_j \frac{\left\|\bm{\Sigma}_{2,j}\right\|_2}{\left\|\mR_j^{\ast}\right\|_2}
\right\},
\end{align*}
where $\bm{\Sigma}_{2,j}$ denotes the truncated singular values at step $j$,
and $q_j$ is the largest index with $\sigma_{q_j}\ge \tau_j$.
\end{theorem}

\begin{corollary}[Interference bound]
Define 
$\mC_t \coloneq \big[\mR_1^\star,\;\mR_2^\star,\;\dots,\;\mR_t^\star\big]\!\in\!\R^{d\times M}$. Let $\bm{\Delta}$ be any future edit with $\left\|\bm{\Delta}\right\|_2\le \Gamma$, where $\Gamma > 0$. Then
\[
\left\|\bm{\Delta}\widetilde\mP_t\vx\right\|
\le \Gamma\left\|\widetilde \mP_t\!-\!\mP_t^\star\right\|_2\,\left\|\vx\right\| \; \forall\,\vx\in \mathrm{span}\!\left(\mC_t\right),
\]
so the worst-case interference is controlled by the cumulative projector approximation error.
\end{corollary}

\begin{remark}[Numerical truncation]
In finite precision, one can stabilize the update toward the "ideal" SVD by discarding tiny singular values due to data noise or rounding errors from forming $\mQ_j$. After truncation, $\mP_{t-1}$ remains a controlled approximation of the ideal projector, with deviation governed by the spectral gap at the truncation index and the discarded tail energy (truncated small singular values)~\citep{wedin1972perturbation,davis1970rotation}. 
\end{remark}  




\section{Experiments and Results}
In this section, we evaluate our approach against several representative model editing methods. The comparison covers both sequential editing performance and the general editing capability of large language models (LLMs). To further validate the design of \methodname, we also conduct a comprehensive ablation study and efficiency analysis.

\subsection{Experimental Setup}

\paragraph{Datasets and Backbone LLMs.}
We perform experiments on two widely used benchmarks: the \textsc{CounterFact}~\citep{ROME} and \textsc{ZsRE} datasets~\citep{Zsre}. To assess model-agnostic applicability, we evaluate on four LLM families of different scales: \texttt{Llama-3} (3B and 8B)~\citep{llama3}, Qwen2.5 (7B)~\citep{qwen-2.5}, GPTJ-6B~\citep{gpt-j} and GPT2-XL~\citep{radford2019language}. 

\paragraph{Baseline Methods and Evaluation Metrics.}
We compare our approach with several established model editing methods, including  \texttt{ROME}~\citep{ROME}, \texttt{MEMIT}~\citep{MEMIT}, and \texttt{AlphaEdit}~\citep{Alphaedit}. Following prior work, we adopt five standard metrics to assess performance: \textit{Efficacy}, \textit{Generalization}, \textit{Specificity}, \textit{Fluency}, and \textit{Consistency}.

\subsection{Factual Editing Results}

\cref{tab:overall_comp} presents a comprehensive comparison of our proposed \methodname{} with prior model editing methods across multiple language models and two benchmark datasets—\textsc{Counterfact} and \textsc{ZsRE} after \textbf{2\textsf{K}} edits. \methodname{} consistently achieves the highest or second-highest performance across nearly all evaluation metrics, including \textit{Efficacy}, \textit{Generalization}, \textit{Specificity}, \textit{Fluency}, and \textit{Consistency}. For the \textsc{Counterfact} benchmark, \methodname~notably surpasses \texttt{AlphaEdit} and other strong baselines such as \texttt{MEMIT} and \texttt{ROME}, achieving higher rewrite success without sacrificing fluency and consistency. 

On \textsc{ZsRE}, \methodname{} further exhibits remarkable generalization and specificity, indicating robust transfer of the edited knowledge across diverse contexts. Across all three backbone models, including \texttt{Llama-3-8B}, \texttt{Qwen2.5-7B-Instruct} and \texttt{Llama-3.2-3B}, our method maintains top-tier efficacy and consistency, underscoring its scalability and reliability in sequential editing scenarios. Overall, these results validate that \methodname{} provides more accurate, consistent, and generalizable edits than existing approaches, establishing a new state of the art in sequential knowledge editing.

\begin{table}[t]
\centering
\resizebox{\linewidth}{!}{
\begin{tabular}{lccccc}
\toprule
\textbf{10K Edits} & \textbf{Eff.$\uparrow$} & \textbf{Gen.$\uparrow$} & \textbf{Spe.$\uparrow$} & \textbf{Flu.$\uparrow$} & \textbf{Consis.$\uparrow$} \\
\midrule
\texttt{MEMIT}        & 49.73 & 49.24 & 51.54 & 389.31 & 3.45  \\
\texttt{ROME}         & 47.57 & 48.45 & 52.52 & 465.02 & 1.83  \\
\texttt{GRACE}        & 7.83  & 10.21 & \textbf{89.20} & \underline{568.49} & 3.61  \\
\texttt{FT}           & \underline{94.04} & \underline{84.12} & 38.15 & 401.57 & \underline{21.35} \\
\texttt{AlphaEdit}   & 66.78 & 58.27 & 51.79 & 489.91 & 4.59  \\
\methodname & \textbf{98.29} & \textbf{91.21} & \underline{63.91} & \textbf{613.88} & \textbf{33.22} \\
\bottomrule
\end{tabular}}
\caption{Performance comparison on the CounterFact benchmark using llama3-8B under 10{,}000 sequential factual edits. Bold denotes the best result and underline denotes the second best. \methodname{} achieves the strongest performance on four of five metrics.}
\label{tab:10k}
\end{table}

We further scale our evaluation to \textbf{10\textsf{K}} edits on \texttt{Llama-3-8B} using a batch size of 100 (\cref{tab:10k}). Under this large-scale editing regime, \methodname{} achieves the strongest performance across all five metrics, surpassing both parametric and non-parametric baselines by substantial margins. These results demonstrate that \methodname{} maintains high effectiveness, generalization, specificity, fluency, and consistency even as the number of edits grows, highlighting its robustness and stability in extended editing scenarios.

\begin{figure}[th!]
  \centering
\includegraphics[width=0.98\linewidth]{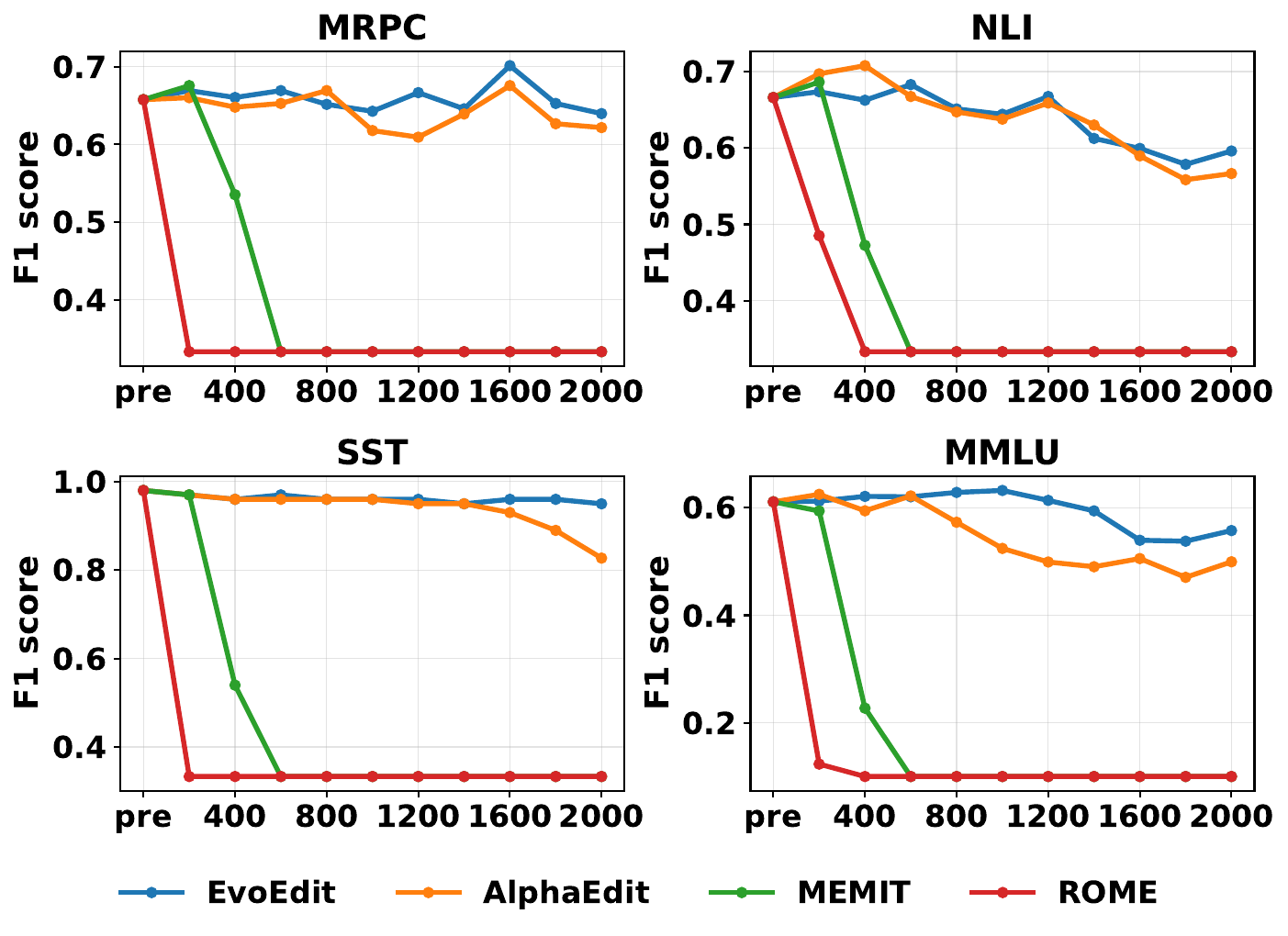}
  \caption{Comparison of \methodname{} with three other baselines (\texttt{ROME, MEMIT, \texttt{AlphaEdit}})  on \texttt{Llama-3-8B} across four evaluation benchmarks (\textsc{MRPC}, \textsc{NLI}, \textsc{SST}, and \textsc{MMLU}). The experiments are conducted with a batch size of 1 over a total of 2000 sequential edits. The $x$-axis represents the cumulative number of edits applied to the model, while the $y$-axis indicates the corresponding F1 score, reflecting the effectiveness of each method in preserving and updating model knowledge over successive edits.}
  \label{fig:glue_four_panel}
\end{figure}
\subsection{General Capability Tests}

\paragraph{Setup.} To evaluate the impact of sequential editing on foundational model performance, we monitor F1 scores across four representative tasks from the GLUE benchmark: \textsc{SST} (sentiment classification), \textsc{MRPC} (paraphrase detection), \textsc{MMLU} (broad knowledge reasoning), and \textsc{NLI} (natural language inference). These benchmarks serve as a vital proxy for the model's linguistic integrity and reasoning stability throughout the editing process. 

\paragraph{Performance.} As illustrated in Figure~\ref{fig:glue_four_panel}, \methodname{} exhibits superior robustness in preserving general capabilities compared to all baseline methods. While \texttt{ROME} and \texttt{MEMIT} suffer from a catastrophic performance collapse—with F1 scores dropping to baseline levels as early as 400 and 800 edits, respectively—\methodname{} maintains high, stable scores throughout the entire 2,000-edit sequence. Even compared to \texttt{AlphaEdit}, which demonstrates a noticeable degradation on \textsc{SST} and \textsc{MRPC} as edits progress, \methodname{} demonstrates a remarkably stable trajectory, particularly evident in the \textsc{SST} and \textsc{MMLU} panels. Ultimately, even at the 2,000-edit mark, \methodname{} preserves a significant portion of its pre-edit performance, confirming that its dynamic projector effectively mitigates the interference that typically cripples standard sequential editing paradigms.

\begin{figure}[th!]
    \centering
    \includegraphics[width=0.99\linewidth]{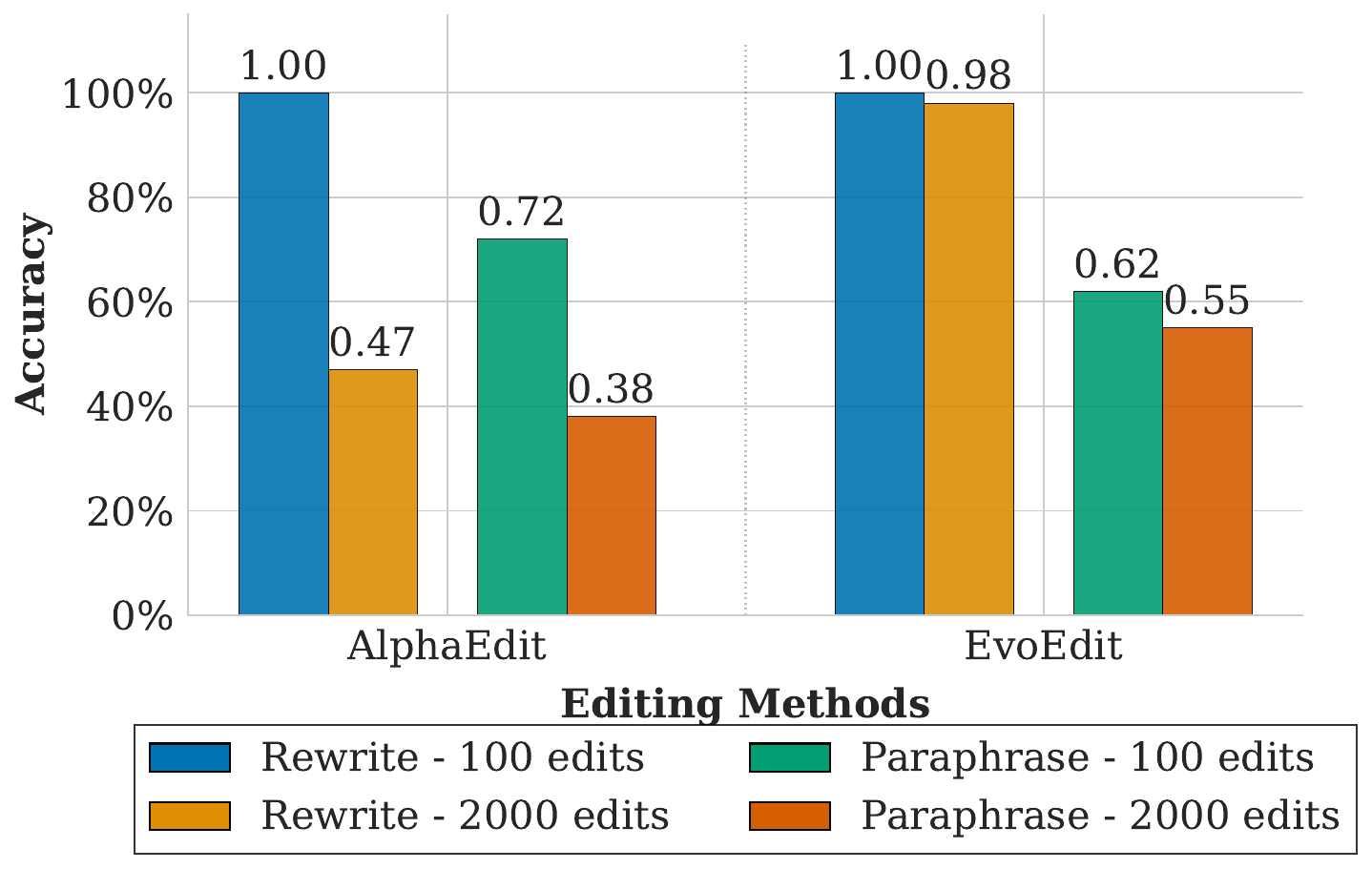}
    \caption{Rewrite and Paraphrase accuracy (\%) on the \emph{first 100} edited facts under two editing methods (\texttt{{AlphaEdit}} and \methodname{}). We report performance at two stages: immediately after the first 100 edits (\textit{100 edits}) and after \textit{2000} total edits. For each stage, bars show accuracy on the \textit{Rewrite} and \textit{Paraphrase} probes. We use a batch size of 1. Evaluation is conducted using \texttt{Llama-3-8B} on \textsc{Counterfact}.}
    \label{fig:drop}
\end{figure}
\section{Analysis and Discussion}
\subsection{Robust Analysis}
\paragraph{Accuracy Drop of Earlier Edits.}
We investigate how well earlier edits are preserved when subsequent edits are added. Specifically, we record the rewrite accuracy and paraphrase accuracy of the first 100 edits at two checkpoints: after 100 (left) and 2000 (right) edits. \cref{fig:drop} shows a substantial performance drop for these early edits in both rewrite and paraphrase accuracy (a decrease of 53\% in rewrite accuracy and 34\% in paraphrase accuracy), whereas the drop under \methodname{} is much smaller than under \texttt{AlphaEdit} (a decrease of  2\% in rewrite accuracy and 7\% in paraphrase accuracy). This serves as an indication that the null-space projector updates are effective in future edits to not interfere with previous ones, suggesting that \methodname{} can better preserve earlier edits.

\subsection{Efficiency Analysis}

\paragraph{Speed-up.} We evaluate editing efficiency of \methodname{} against \texttt{AlphaEdit} on an NVIDIA H100 (80GB) using a fixed sequence of 500 edits to ensure a fair comparison. As summarized in \cref{tab:runtime_batch_compact}, \methodname{} substantially reduces per-edit latency and delivers significant speed-ups across different batch sizes (BS). We specifically benchmark two critical components: the time required to compute the update matrix (\emph{Solve}) and the time required to re-compute the projection matrix at each step using the update (\emph{Proj}). In detail, the $O\!\left(d^3\right)$ solver used by \texttt{AlphaEdit} is replaced in \methodname{} by a smaller $k_t$-sized inner system, minimizing the \emph{Solve} cost. While \texttt{AlphaEdit} does not update projection matrices, \methodname{} still achieves substantial gains in total wall-clock time, ranging from 1.98$\times$ to 3.53$\times$ faster depending on the model and batch size. 

\paragraph{Memory consumption.} GPU memory usage over 1000 edits on \texttt{Llama-3-8B} shows that \methodname{} delivers substantial efficiency gains, reducing peak allocated and peak reserved memory by up to 14\% and 15\% respectively (\cref{tab:memory_usage}), indicating high-quality edits simultaneous to meaningful speedups and a leaner memory footprint.


\renewcommand{\spdup}[1]{\tt\textcolor{green!40!black}{\tiny(x#1)}}

\providecommand{\sol}[1]{\textcolor{red!70!black}{#1}}         
\providecommand{\prj}[1]{\textcolor{cyan!60!black}{#1}}        
\providecommand{\spdup}[1]{\textcolor{green!40!black}{(x#1)}}  

\begin{table}[t]
  \centering
  \scriptsize
  \setlength{\tabcolsep}{5pt}
  \resizebox{\linewidth}{!}{
  \begin{tabular}{@{}lrrrrrr@{}}
    \toprule
    \multirow{2.5}{*}{\textbf{Model / BS}}& \multicolumn{2}{c}{\texttt{{AlphaEdit}} (s)} & \multicolumn{3}{c}{\methodname{} (s)} \\
    \cmidrule(l){2-3}\cmidrule(l){4-6}
    & Solve $\downarrow$ & Total $\downarrow$
                        & Solve $\downarrow$ & Proj $\downarrow$ & \multicolumn{1}{c}{Total $\downarrow$} \\
    \midrule {\textbf{\texttt{Llama-3-8B}}} \\
    \quad BS=1
     & \sol{1313.6}    & 1313.6 & \sol{1.4} & \prj{661.0} & {662.4 \spdup{1.98}} \\
      \quad BS=10  & \sol{224.5}     &  224.5 & \sol{0.3} & \prj{141.7}  & {142.0  \spdup{1.58}} \\
      \quad BS=100 & \sol{22.0}      &   22.0 & \sol{0.1} & \prj{8.8}    & {8.9    \spdup{2.47}} \\
    \midrule
    \textbf{Qwen2.5-7B} \\
    \quad BS=1 & \sol{3251.3}    & 3251.3 & \sol{2.3} & \prj{1491.4} & {1493.8 \spdup{2.18}} \\
      \quad BS=10  & \sol{461.6}     &  461.6 & \sol{0.4} & \prj{169.0}  & {169.4  \spdup{2.72}} \\
      \quad BS=100 & \sol{39.9}      &   39.9 & \sol{0.1} & \prj{11.2}   & {11.3   \spdup{3.53}} \\
    \bottomrule
  \end{tabular}
  }
  \caption{Runtime on the MCF dataset across batch sizes (BS) for \texttt{Llama-3-8B} and \texttt{Qwen2.5-7B-Instruct} for 500 total edits. \textit{Solve}: time to compute the update matrix; \textit{Proj}: time to update the projector; \textit{Total}: overall runtime in seconds ($\downarrow$ better).}
  \label{tab:runtime_batch_compact}
\end{table}


\begin{table}[t]
\centering
\small
\resizebox{\linewidth}{!}{
\begin{tabular}{lcc}
\toprule
\textbf{Method} & \textbf{Peak Alloc. (GB)} & \textbf{Peak Reserved (GB)} \\
\midrule
\texttt{MEMIT}     & 36.87 & 38.45 \\
\texttt{AlphaEdit} & 34.79 & 35.36 \\
\methodname{}   & \textbf{31.73} \,(\textbf{-14\%}) & \textbf{32.74} \,(\textbf{-15\%}) \\
\bottomrule
\end{tabular}}
\caption{GPU memory usage over 1000 edits on \texttt{Llama-3-8B}. \methodname{} reduces peak allocated and reserved memory compared to prior methods.}
\label{tab:memory_usage}
\end{table}

\section{Conclusion}

We propose \methodname, a scalable model editing method that exploits the model’s null space to enable continual edits. By addressing the limitations of fixed null-space projectors in prior work, \methodname{} mitigates interference. We further provide a numerically stable formulation that reduces editing complexity from cubic to quadratic in the input dimension. Experiments on modern LLMs such as \texttt{Llama} and \texttt{Qwen} validate our theory, showing improved editing performance across benchmarks while achieving substantial runtime speedups.

\section{Limitations and Ethical Considerations}

The primary limitation of our work remains the finite number of models and datasets on which our method is tested. Furthermore, there are specific scenarios where the tested datasets do not cover, for example, controlling for the relatedness between sequentially edited facts.

As our work only looks at sequential model editing, we do not foresee or anticipate any specific ethical considerations to be acknowledged. However, like all model-editing methods, there are scenarios where such techniques could be used to apply undesirable knowledge or traits within models, which can be worth future discussion.

\bibliography{custom}
\appendix
\onecolumn

\section{Proofs}
\label{sec:proofs}

\subsection{Proof of \cref{thm:equivalence} (Exact equivalence without truncation)}

\begin{proposition}
Let $\mK_0\!\in\!\R^{d\times m}$ and define
\[
\mP_0 = \mI_d - \mK_0 \left(\mK_0^\top \mK_0\right)^{-1} \mK_0^\top.
\]
Then $\mP_0$ is the orthogonal projector onto $\mathrm{Range}\!\left(\mK_0\right)^\perp$ and $\mathrm{Range}\!\left(\mP_0\right)=\mathrm{Range}\!\left(\mK_0\right)^\perp$.
\end{proposition}

\begin{proof}
Set $\mM=(\mK_0^\top\mK_0)^{-1}$. Since $\mK_0^\top \mK_0$ is symmetric, so is $\mM$. 
Then it follows,
\[
\left(\mK_0 \mM \mK_0^\top\right)^2
= \mK_0 \mM \left(\mK_0^\top \mK_0\right) \mM \mK_0^\top
= \mK_0 \mM \mK_0^\top,
\]
and $\left(\mK_0 \mM \mK_0^\top\right)^\top=\mK_0 \mM \mK_0^\top$, 
so $\mK_0 \mM \mK_0^\top$ is the orthogonal projector onto $\mathrm{Range}\!\left(\mK_0\right)$.  
Therefore $\mP_0=\mI_d-\mK_0\mM\mK_0^\top$ is symmetric and idempotent.\par

\noindent To identify its range, note that for any $\vx$,
\[
\mK_0^\top \mP_0 \vx
= \mK_0^\top \vx - (\mK_0^\top \mK_0)\mM \mK_0^\top \vx
= \mK_0^\top \vx - \mI_d(\mK_0^\top \vx)
= 0,
\]
so $\mathrm{Range}\!\left(\mP_0\right)\subseteq \mathrm{Null}\!\left(\mK_0^\top\right)$. Conversely, if $\vy\in \mathrm{Null}\!\left(\mK_0^\top\right)$ then
$\mP_0 \vy=\vy$, hence $\vy\in\mathrm{Range}\!\left(\mP_0\right)$. Thus
\[
\mathrm{Range}\!\left(\mP_0\right)=\mathrm{Null}\!\left(\mK_0^\top\right)=\mathrm{Range}\!\left(\mK_0\right)^\perp.
\]
\end{proof}

\begin{proposition} \label{le:QP}
If $\mP_{j-1}$ is idempotent and $\mQ_j$ is an orthonormal basis of $\mathrm{Range}\!\left(\mP_{j-1}\mK_j\right)$,
then 
$\mP_{j-1}\mQ_j = \mQ_j$ and $\mQ_j^\top \mP_{j-1} = \mQ_j^\top$.
\end{proposition}

\begin{proof}
Each column $\vq$ of $\mQ_j$ lies in $\mathrm{Range}\!\left(\mR_j\right)\subseteq \mathrm{range}\!\left(\mP_{j-1}\right)$.  
Hence $\vq=\mP_{j-1}\vz$ for some $\vz$. Then
\[
\mP_{j-1}\vq = \mP_{j-1}\!\left(\mP_{j-1}\vz\right) = \mP_{j-1}^2 \vz = \mP_{j-1}\vz = \vq,
\]
since $\mP_{j-1}$ is idempotent. This proves $\mP_{j-1}\mQ_j=\mQ_j$.  
The second identity follows by transposition.
\end{proof}

\begin{proposition}\label{le:QP2}
Let $\mP_{j-1}$ and $\mQ_j$ be defined as in Proposition \ref{le:QP}.
Then $\mP_j \triangleq \mP_{j-1} - \mQ_j \mQ_j^\top$ is also an orthogonal projector.
\end{proposition}

\begin{proof} 
It is obvious that $\mP_j$ is symmetric. 
We just need to show it is idempotent.
\[
\begin{aligned}
\mP_j^2 &= \left(\mP_{j-1}-\mQ_j\mQ_j^\top\right)^2 \\
&= \mP_{j-1}^2 - \mP_{j-1}\mQ_j\mQ_j^\top - \mQ_j\mQ_j^\top \mP_{j-1} + \mQ_j\mQ_j^\top \mQ_j\mQ_j^\top.
\end{aligned}
\]
Using $\mP_{j-1}^2=\mP_{j-1}$, $\mP_{j-1}\mQ_j=\mQ_j$, $\mQ_j^\top \mP_{j-1}=\mQ_j^\top$ (by Proposition A.2.), and $\mQ_j^\top \mQ_j=\mI$, we obtain
\[
\mP_j^2 = \mP_{j-1} - \mQ_j\mQ_j^\top = \mP_j.
\]
Thus $\mP_j$ is symmetric and idempotent, hence an orthogonal projector.
\end{proof}

\begin{lemma}[Projector difference for nested subspaces]
\label{lem:proj-diff}
Let $S,U\subseteq\R^d$ be linear subspaces with $\mU\subseteq S$.
Let $\mP_S$ and $\mP_U$ be the orthogonal projectors onto $S$ and $\mU$, respectively.
Then
\[
\mP\coloneq\mP_S - \mP_U
\]
is the orthogonal projector onto the subspace $S\cap \mU^\perp$; in particular,
\[
\mathrm{range}\!\left(\mP\right)=S\cap \mU^\perp
\quad\text{and}\quad
\mathrm{Null}\!\left(\mP\right)=\mU\ \oplus\ S^\perp.
\]
\end{lemma}

\begin{proof}
Since $\mU\subseteq S$, $\mP_S\mP_U=\mP_U \mP_S=\mP_U$. Hence
\[
\mP^\top=(\mP_S-\mP_U)^\top=\mP_S-\mP_U=\mP,\qquad
\mP^2=(\mP_S-\mP_U)^2=\mP_S-\mP_U.
\]
Thus $\mP$ is an orthogonal projector. For any $x$, write $x=s+s_\perp$ with $s=\mP_Sx\in S$ and $s_\perp\in S^\perp$, and then decompose $s=\mU+w$ with $\mU=\mP_U s\in \mU$ and $w\in S\cap \mU^\perp$. We get
\[
\mP x=(\mP_S-\mP_U)(s+s_\perp)=s-u=w\in S\cap \mU^\perp,
\]
so $\mathrm{range}\!\left(\mP\right)\subseteq S\cap \mU^\perp$. Conversely, for any $w\in S\cap \mU^\perp$, $(\mP_S-\mP_U)w=w$, hence $w\in\mathrm{range}\!\left(\mP\right)$ and $\mathrm{range}\!\left(\mP\right)=S\cap \mU^\perp$. Taking orthogonal complements yields $\mathrm{Null}\!\left(\mP\right)=\mU\oplus S^\perp$.
\end{proof}

\begin{corollary}[Instantiation for our update]
\label{cor:instantiation}
Let $S\coloneq\mathrm{range}\!\left(\mP_{t-1}\right)$, $\mU\coloneq\mathrm{span}\!\left(\mR_t\right)=\mathrm{Range}\!\left(\mQ_t\right)$ with $\mR_t=\mP_{t-1}\mK_t$ and $\mathrm{Range}\!\left(\mQ_t\right)=\mathrm{Range}\!\left(\mR_t\right)\subseteq S$.
Then, for $\mP_t\coloneq\mP_{t-1}-\mQ_t\mQ_t^\top$,
\[
\mathrm{range}\!\left(\mP_t\right))=\mathrm{range}\!\left(\mP_{t-1}\right)\cap \mathrm{span}\!\left(\mR_t\right)^\perp,
\qquad
\mathrm{Null}\!\left(\mP_t\right))=\mathrm{Null}\!\left(\mP_{t-1}\right)\ \oplus\ \mathrm{span}\!\left(\mR_t\right).
\]
\end{corollary}

\begin{proof}
Apply Lemma~\ref{lem:proj-diff} with $\mP_S=\mP_{t-1}$ and $\mP_U=\mQ_t\mQ_t^\top$.
\end{proof}

\begin{theorem}[Exact equivalence without truncation]
If no truncation is applied when forming $\mQ_j$, then for all $t\ge 0$,
\[
\mathrm{Null}\!\left(\mP_t\right) = \mathrm{span}\!\left(\mK_0,\dots,\mK_t\right),
\qquad
\mP_t=\mP_t^\star.
\]
\end{theorem}

\begin{proof}
\textit{Base case ($t=0$).}
By Proposition A.1, $\mP_0$ is the orthogonal projector onto $\mathrm{span}\!\left(\mK_0\right)^\perp$, hence
$\mathrm{Null}\!\left(\mP_0\right)=\mathrm{span}\!\left(\mK_0\right)$.

\textit{Induction step.}
Assume $\mathrm{Null}\!\left(\mP_{t-1}\right)=\mathrm{span}\!\left(\mK_0,\dots,\mK_{t-1}\right)$ and assume $\mP_{t-1}$ is an orthogonal projector.
Write
\[
\mK_t=\left(\mI-\mP_{t-1}\right)\mK_t+\mP_{t-1}\mK_t \;=:\; \mB_t+\mR_t,
\]
so that $\mathrm{Range}\!\left(\mB_t\right)\subseteq \mathrm{Null}\!\left(\mP_{t-1}\right)$ and
    $\mathrm{Range}\!\left(\mR_t\right)\subseteq \mathrm{range}\!\left(\mP_{t-1}\right)$.
Since $\mathrm{Null}\!\left(\mP_{t-1}\right)\perp \mathrm{range}\!\left(\mP_{t-1}\right)$, we have the orthogonal direct sum
\begin{align*}
\mathrm{span}\!\left(\mK_0,\dots,\mK_t\right)
&= \mathrm{span}\!\left(\mK_0,\dots,\mK_{t-1},\,\mB_t+\mR_t\right) \\
&= \mathrm{span}\!\left(\mK_0,\dots,\mK_{t-1}\right)\ \oplus\ \mathrm{span}\!\left(\mR_t\right) \\
&= \mathrm{Null}\!\left(\mP_{t-1}\right)\ \oplus\ \mathrm{span}\!\left(\mR_t\right).
\end{align*}
On the other hand, since both $\mP_{t-1}$ and $\mQ_t\mQ_t^\top$ are orthogonal projectors, by  Proposition~\ref{le:QP2} and Lemma~\ref{lem:proj-diff}, $\mP_t=\mP_{t-1}-\mQ_t\mQ_t^\top$ is also an orthogonal projector with $\mathrm{Range}\!\left(\mQ_t\right)=\mathrm{span}\!\left(\mR_t\right)\subseteq \mathrm{range}\!\left(\mP_{t-1}\right)$. Hence, by Corollary~\ref{cor:instantiation},
\begin{align*}
\mathrm{Null}\!\left(\mP_t\right)
&= \mathrm{Null}\!\left(\mP_{t-1}\right)\ \oplus\ \mathrm{span}\!\left(\mR_t\right) \\
&= \mathrm{span}\!\left(\mK_0,\dots,\mK_t\right).
\end{align*}
Finally, since $\mP_t$ is an orthogonal projector and its null space equals $\mathrm{span}\!\left(\mK_0,\dots,\mK_t\right)$, it must coincide with the unique projector onto $\mathrm{span}\!\left(\mK_0,\dots,\mK_t\right)^\perp$, i.e., $\mP_t=\mP_t^\star$.
\end{proof}

\subsection{Proof of \cref{thm:global_error} (Global error bound with truncation)}

We first give a lemma that separates the
retained-part (geometric) deviation from the discarded-part (tail-energy) error.
This avoids comparing subspaces of different dimensions by only applying
$\sin\Theta$ to $q$-dimensional subspaces. \par
\noindent \textbf{Notation} A superscript ${}^\ast$ denotes the \emph{ideal, no-truncation} quantity. 
In particular, $\mP_{j-1}^\ast$ is the orthogonal projector onto 
$\mathrm{span}\!\left(\mK_0,\ldots,\mK_{j-1}\right)^\perp$, and 
$\mR_j^\ast=\mP_{j-1}^\ast\mK_j$. 
By contrast, $\mP_{j-1}$ is the algorithmic projector after $j{-}1$ steps with truncation, and 
$\mR_j=\mP_{j-1}\mK_j=\mR_j^\ast+\mE_j$ with $\mE_j=\left(\mP_{j-1}-\mP_{j-1}^\ast\right)\mK_j$.
For two subspaces $\mathcal U,\mathcal V\subset\R^d$ with orthonormal bases $U,V$ and
orthogonal projectors $P_{\mathcal U}=UU^\top$, $P_{\mathcal V}=VV^\top$, let
$\Theta(\mathcal U,\mathcal V)=(\theta_1,\ldots,\theta_q)$ denote the vector of principal angles
(with $q=\min\{\dim\mathcal U,\dim\mathcal V\}$). We write
\[
\|\sin\Theta(\mathcal U,\mathcal V)\|\ :=\ \max_{i}\sin\theta_i,
\]
and recall the identity
\begin{equation}
\label{eq:sin-theta-proj}
\|P_{\mathcal U}-P_{\mathcal V}\|_2\ =\ \|\sin\Theta(\mathcal U,\mathcal V)\|.
\end{equation}
\begin{lemma}(Refined, gap--tail form)
\label{lem:anglebound-refined}
Let $\mR_j^\ast = \mP_{j-1}^\ast \mK_j$ with SVD
$\mR_j^\ast = \mU\bm{\Sigma}\mV^\top$, where the nonzero singular values are
$\sigma_1\ge\dots\ge\sigma_r>0$ and $\mU=[\mU_1\ \mU_2]$ with $\mU_1\in\R^{d\times q}$.
Given a threshold $\tau_j>0$, let $q=\max\{i:\sigma_i\ge\tau_j\}$, and construct
$\widehat \mU_1$ by applying the same truncation rule to
$\mR_j=\mP_{j-1}\mK_j=\mR_j^\ast+\mE_j$, where
$\mE_j=(\mP_{j-1}-\mP_{j-1}^\ast)\mK_j$.
Define the projectors $\mP_1=\mU_1\mU_1^\top$ and $\widehat \mP_1=\widehat \mU_1\widehat \mU_1^\top$,
and set $\mP^\star=\mU\mU^\top$.

\begin{enumerate}
\item[(i)] (\textbf{No truncation}) If $\tau_j\le\sigma_r$ then $q=r$, $\mU_1=\mU$, and
$\widehat \mU_1$ has the same dimension; hence
    $\sin\Theta\left(\mathrm{span}\!\left(\mU_1\right),\mathrm{span}\!\left(\widehat \mU_1\right)\right)=0$ whenever $\mE_j=0$,
and in general
\[
\left\|\sin\Theta\left(\mathrm{span}\!\left(\mU_1\right),\mathrm{span}\!\left(\widehat \mU_1\right)\right)\right\|
\ \le\ \frac{\left\|\mE_j\right\|_2}{\sigma_r}\,.
\]
\item[(ii)] (\textbf{With truncation}) If $\tau_j\in(\sigma_{q+1},\sigma_q)$ for some $q<r$, let the spectral gap be $\gamma_j=\sigma_q-\sigma_{q+1}>0$. Then
\begin{align}
\left\|\sin\Theta\left(\mathrm{span}\!\left(\mU_1\right),\mathrm{span}\!\left(\widehat \mU_1\right)\right)\right\|
&\ \le\ \frac{\left\|\mE_j\right\|_2}{\gamma_j},
\label{eq:wedin}\\[4pt]
\left\| \left(\mP_1-\mP^\star\right)\,\mR_j^\ast \right\|_F^2
&\ =\ \sum_{i>q}\sigma_i^2. \label{eq:tail}
\end{align}
\end{enumerate}
\end{lemma}

\begin{proof}
The bound \cref{eq:wedin} is a standard Davis--Kahan--Wedin type result
for $q$-dimensional left singular subspaces under additive perturbation
$E_j$, with denominator given by the gap between the $q$-th and
$(q{+}1)$-th singular values of $\mR_j^\ast$.
For \cref{eq:tail}, write
\[
(\mP_1-\mP^\star)\,\mR_j^\ast
= (\mU_1\mU_1^\top-\mU\mU^\top)\,\mU\bm{\Sigma}\mV^\top
= -\mU_2\bm{\Sigma}_2 \mV^\top,
\]
where $\bm{\Sigma}_2=\mathrm{diag}\!\left(\sigma_{q+1},\dots,\sigma_r\right)$.
Taking Frobenius norms yields $\left\| \left(\mP_1-\mP^\star\right)\,\mR_j^\ast \right\|_F^2
=\left\|\bm{\Sigma}_2\right\|_F^2=\sum_{i>q}\sigma_i^2$.
\end{proof}

\noindent We next prove the theorem by using the lemma. We bound the one-step \emph{action} of the projector error on the signal
$\mR_j^\ast$. Using $\widetilde \mP_j=\mP_{j-1}-\widehat \mP_1$ and $\mP_j^\ast=\mP_{j-1}^\ast-\mP^\ast$, we have the exact decomposition
\[
\widetilde \mP_j - \mP_j^\ast
= (\mP_{j-1}-\mP_{j-1}^\ast) \;+\; \left(\mP^\ast-\mP_1\right) \;+\; \left(\mP_1-\widehat \mP_1\right).
\]
Hence, by the triangle inequality and sub-multiplicativity of operator/Frobenius norms,
\begin{equation}
\label{eq:onestep-action-3term}
\left\|\left(\widetilde \mP_j - \mP_j^\ast\right)\,\mR_j^\ast\right\|_F
\ \le\ \underbrace{\left\|\left(\mP_{j-1}-\mP_{j-1}^\ast\right)\,\mR_j^\ast\right\|_F}_{\text{carry-over}}
\;+\;\underbrace{\left\|\left(\mP^\ast-\mP_1\right)\,\mR_j^\ast\right\|_F}_{\text{tail}}
\;+\;\underbrace{\left\|\left(\mP_1-\widehat \mP_1\right)\,\mR_j^\ast\right\|_F}_{\text{geometry}}.
\end{equation}
For the last two terms, by \eqref{eq:sin-theta-proj} and \cref{eq:wedin},
\[
\|(\mP_1-\widehat \mP_1)\,\mR_j^\ast\|_F
\ \le\ \|\mP_1-\widehat \mP_1\|_2\,\|\mR_j^\ast\|_F
\ =\ \|\sin\Theta(\mathrm{span}(\mU_1),\mathrm{span}(\widehat\mU_1))\|\,\|\mR_j^\ast\|_F
\ \le\ \frac{\|\mE_j\|_2}{\gamma_j}\,\|\mR_j^\ast\|_F.
\] Therefore,
\begin{equation}
\label{eq:onestep-gap-tail-3}
\left\|\left(\widetilde \mP_j - \mP_j^\ast\right)\,\mR_j^\ast\right\|_F
\ \le\ \left\|\left(\mP_{j-1}-\mP_{j-1}^\ast\right)\,\mR_j^\ast\right\|_F
\;+\;\frac{\left\|\mE_j\right\|_2}{\gamma_j}\,\left\|\mR_j^\ast\right\|_F
\;+\;\left\|\bm{\Sigma}_2\right\|_F.
\end{equation}

\noindent \textit{Absorbing the carry-over term.}
By the induction hypothesis at step $j{-}1$ we already have
\(
\|\mP_{j-1}-\mP_{j-1}^\ast\|_2 \le B_{j-1},
\)
where $B_{j-1}$ denotes the cumulative gap--tail bound up to step $j{-}1$. Hence
\[
\left\|\left(\mP_{j-1}-\mP_{j-1}^\ast\right)\,\mR_j^\ast\right\|_F
\ \le\ \|\mP_{j-1}-\mP_{j-1}^\ast\|_2\,\|\mR_j^\ast\|_F
\ \le\ B_{j-1}\,\|\mR_j^\ast\|_F.
\]

where $B_{j-1} = \frac{1}{\sigma_{\min}\!\big(\mC_{j-1}\big)}
\sum_{i=1}^{\,j-1}\!\left(
\frac{\|\mE_i\|_2}{\gamma_i}\,\|\mR_i^\ast\|_F \;+\; \|\bm{\Sigma}_{2,i}\|_F
\right)$. By induction on $j$ (unwinding the same bound for steps $1,\dots,j-1$), this carry-over factor is controlled by the already accumulated gap–tail terms up to step $j{-}1$. For exposition we subsume it into the geometric factor and keep the displayed right-hand side in the simple ``gap–tail" form used below; the full three-term recurrence \cref{eq:onestep-gap-tail-3} only strengthens the final bound.

\noindent Using the above, we obtain the one-step \emph{gap–tail} bound
\begin{equation}
\label{eq:onestep-gap-tail}
\left\|\left(\widetilde \mP_j - \mP_j^\ast\right)\,\mR_j^\ast\right\|_F
\ \le\ \frac{\left\|\mE_j\right\|_2}{\gamma_j}\,\left\|\mR_j^\ast\right\|_F\;+\;\left\|\bm{\Sigma}_2\right\|_F.
\end{equation}

\noindent We now pass to the global bound. Let $\mC_t=[\mR_1^\ast,\dots,\mR_t^\ast]\in\R^{d\times M}$,
and recall $\mathcal C_t=\mathrm{span}\!\left(\mC_t\right)$.
Define the \emph{restricted operator norm} on $\mathcal C_t$:
\[
\left\|\mA\right\|_{2\to2}^{(\mathcal C_t)} \coloneq \sup_{x\in\mathcal C_t,\ x\neq 0}\frac{\left\|\mA\vx\right\|_2}{\left\|\vx\right\|_2}.
\]
Using induction on $t$ and the fact that at step $j$ the update only acts through the rank-$q_j$ projector change on $\mR_j^\ast$, we obtain
\begin{align}
\left\|\left(\widetilde \mP_t\!-\!\mP_t^\ast\right)\,\mC_t\right\|_F
&\ \le\ \sum_{j=1}^t \left\|\left(\widetilde \mP_j - \mP_j^\ast\right)\,\mR_j^\ast\right\|_F \nonumber\\
&\ \le\ \sum_{j=1}^t\left(
\frac{\left\|\mE_j\right\|_2}{\gamma_j}\,\left\|\mR_j^\ast\right\|_F\;+\;\left\|\bm{\Sigma}_{2,j}\right\|_F\right),
\label{eq:global-frob}
\end{align}
where $\bm{\Sigma}_{2,j}$ is the discarded block at step $j$.

\noindent Finally, for any $\vx\in\mathcal C_t$ write $\vx=\mC_t \bm{\alpha}$ with
$\bm{\alpha}\in\R^{M}$. Then
\[
\left\|\left(\widetilde \mP_t\!-\!\mP_t^\ast\right)\,\vx\right\|_2
\ \le\ \frac{\left\|\left(\widetilde \mP_t\!-\!\mP_t^\ast\right)\,\mC_t\right\|_F}{\sigma_{\min}\!\left(\mC_t\right)}\,\left\|\vx\right\|_2,
\]
where $\sigma_{\min}\!\left(\mC_t\right)$ is the smallest nonzero singular value of $\mC_t$.
Hence
\begin{equation}
\label{eq:restricted-operator}
\| \widetilde \mP_t\!-\!\mP_t^\ast \|_{2\to2}^{(\mathcal C_t)}
\ \le\ \frac{1}{\sigma_{\min}\!\left(\mC_t\right)}\,
\sum_{j=1}^t\left(
\frac{\left\|\mE_j\right\|_2}{\gamma_j}\,\|\mR_j^\ast\|_F\;+\;\left\|\bm{\Sigma}_{2,j}\right\|_F\right).
\end{equation}
Since both $\widetilde \mP_t$ and $\mP_t^\ast$ are orthogonal projectors, their
unrestricted spectral-norm difference is at most $1$, so we may write the
final global bound as
\[
\|\widetilde \mP_t\!-\!\mP_t^\ast\|_2
\ \le\ \min\left\{\,1,\;
\frac{1}{\sigma_{\min}\!\left(\mC_t\right)}\,
\sum_{j=1}^t\left(
\frac{\left\|\mE_j\right\|_2}{\gamma_j}\,\|\mR_j^\ast\|_F+\left\|\bm{\Sigma}_{2,j}\right\|_F\right)
\right\}.
\]
This completes the proof of \cref{thm:global_error}.

\subsection{Proof of \cref{cor:interference} (Interference bound)}

\paragraph{Notation recalled.}
For $t\ge 1$, define the (block) matrix
\[
\mC_t \;:=\; \big[\mR_1^\star,\;\mR_2^\star,\;\dots,\;\mR_t^\star\big]\!\in\!\R^{d\times M},
\quad\text{so that}\quad \mathrm{span}(\mC_t)=\mathrm{span}\!\big(\mR_1^\star,\dots,\mR_t^\star\big).
\]
The ideal projector $\mP_t^\star$ is the orthogonal projector onto the orthogonal complement
of $\mathrm{span}(\mC_t)$, i.e.,
\[
\mP_t^\star \;=\; \mathrm{span}(\mC_t)^\perp
\quad\Longrightarrow\quad
\mP_t^\star x \;=\; 0,\quad \forall\,x\in \mathrm{span}(\mC_t).
\]
Let $\widetilde \mP_t$ be the truncated projector constructed in the main text.
For matrices, $\|\cdot\|_2$ denotes the spectral norm; for vectors, $\|\cdot\|$ is the Euclidean norm.

\begin{corollary}[Interference bound]\label{cor:interference}
Let $\bm{\Delta}\in\R^{d\times d}$ be any (future) linear edit with $\|\bm{\Delta}\|_2\le \Gamma$.
Then for every $\vx\in \mathrm{span}(\mC_t)$,
\[
\big\|\bm{\Delta}\,\widetilde \mP_t\,\vx\big\|
\;\le\; \Gamma\,\big\|\widetilde \mP_t\!-\!\mP_t^\star\big\|_2\,\|\vx\|.
\]
In particular, the worst-case interference on $\mathrm{span}(\mC_t)$ is controlled by the projector
approximation error $\|\widetilde \mP_t\!-\!\mP_t^\star\|_2$.
\end{corollary}

\begin{proof}
Fix any $\vx\in \mathrm{span}(\mC_t)$. By definition of $\mP_t^\star$,
we have $\mP_t^\star \vx = 0$. Therefore
\[
\widetilde \mP_t\,\vx \;=\; \big(\widetilde \mP_t\!-\!\mP_t^\star\big)\vx.
\]
Applying $\bm{\Delta}$ and taking norms, we use submultiplicativity of the operator norm:
\[
\big\|\bm{\Delta}\,\widetilde \mP_t\,\vx\big\|
\;=\; \big\|\bm{\Delta}\,(\widetilde \mP_t\!-\!\mP_t^\star)\,\vx\big\|
\;\le\; \|\bm{\Delta}\|_2\,\|\widetilde \mP_t\!-\!\mP_t^\star\|_2\,\|\vx\|
\;\le\; \Gamma\,\|\widetilde \mP_t\!-\!\mP_t^\star\|_2\,\|\vx\|.
\]
This proves the claim.
\end{proof}

\subsection{Proof of \cref{eq:solution1} (Solution with heavy matrix inversion.)}
\label{sec: solution-proofs}
\begin{proof}
    Given the aligned projector matrix $\mP_{t-1}$, since it is an orthogonal projection, we have $\mP = \mP^\top$ and $\mP^2 = \mP$. The sequential editing objective:
    \begin{align}
        L = \big(\|(\mW_{t-1}+\bm{\Delta}_t\mP_{t-1})\mK_t-\mV_t\|^2 + \|\bm{\Delta}_t\mP_{t-1}\|^2\big).
    \end{align}
    We define $\mR = \mV_t -\mW_{t-1}\mK_t$, setting the matrix derivative $\nabla_{\bm{\Delta}}L = 0$ yields:
    \begin{align}
        (\bm{\Delta}_t\mP_{t-1}\mK_t -\mR)\mK^\top_t\mP_{t-1}^\top + \bm{\Delta}_t\mP\mP^\top = \bm{0}.
    \end{align}
Factorize $\bm{\Delta}_t\mP_{t-1}$ and use $\mP = \mP^\top$ and $\mP^2 = \mP$, we obtain:
\begin{align}
    \bm{\Delta}_t\mP_{t-1}(\mK_t\mK_t^\top\mP_{t-1} + \mI) = \mR\mK_t^\top\mP_{t-1}.
\end{align}
We then get the final closed-form solution if the inversion of the bracketed term.
\begin{align}
    \bm{\Delta}_{\methodname}  = \bm{\Delta}_t\mP_{t-1} = \mR\mK_t^\top\mP_{t-1}\big(\mK_t\mK_t^\top\mP_{t-1} + \mI\big)^{-1}.
\end{align}
\end{proof}
\subsection{Proof of \cref{eq:solution2} (Solution via Woodbury-identity-matrix.)}

\begin{proof}
Let
\[
\mK_t\in\mathbb R^{d\times m},\qquad \mP_{t-1}\in\mathbb R^{d\times d},\qquad \mR\in\mathbb R^{d\times r}.
\]
Start from the given expression
\[
\bm{\Delta}_t\mP_{t-1}
= \mR\,\mK_t^{\top}\mP_{t-1}\Big(\mK_t \mK_t^{\top}\mP_{t-1} + I_d\Big)^{-1}.
\]

We use the standard matrix identity (a rearrangement closely related to Woodbury):
\begin{equation}\label{eq:AB-identity}
\big(\mI_d + \mA \mB\big)^{-1} \mA = \mA \big(\mI_m + \mB \mA\big)^{-1}
\qquad\text{for }\mA\in\mathbb R^{d\times m},\; \mB\in\mathbb R^{m\times d},
\end{equation}
which is verified by multiplying both sides on the left by $(\mI_d+\mA\mB)$ and on the right by $(\mI_m+\mB\mA)$ (or derived from the Woodbury identity).

Take
\[
\mA = \mK_t,\qquad \mB = \mK_t^{\top}\mP_{t-1}.
\]
Then $\mA\mB = \mK_t \mK_t^{\top}\mP_{t-1}$ and $\mB\mA = \mK_t^{\top}\mP_{t-1}\mK_t$. Applying \cref{eq:AB-identity} gives
\[
\big(\mI_d + \mK_t \mK_t^{\top}\mP_{t-1}\big)^{-1} \mK_t
= \mK_t \big(\mI_m + \mK_t^{\top}\mP_{t-1}\mK_t\big)^{-1}.
\]

Multiply the last identity on the left by $\mK_t^{\top}\mP_{t-1}$ (or equivalently take transposes and rearrange) to obtain
\begin{equation}\label{eq:swap}
\mK_t^{\top}\mP_{t-1}\big(\mI_d + \mK_t \mK_t^{\top}\mP_{t-1}\big)^{-1}
= \big(\mI_m + \mK_t^{\top}\mP_{t-1}\mK_t\big)^{-1} \mK_t^{\top}\mP_{t-1}.
\end{equation}

Substitute \cref{eq:swap} into the right-hand side of (1):
\[
\begin{aligned}
\bm{\Delta}_t\mP_{t-1}
&= \mR\;\Big[\,\mK_t^{\top}\mP_{t-1}\big(\mI_d + \mK_t \mK_t^{\top}\mP_{t-1}\big)^{-1}\Big]\\
&= \mR\;\Big[\,\big(\mI_m + \mK_t^{\top}\mP_{t-1}\mK_t\big)^{-1} \mK_t^{\top}\mP_{t-1}\Big]\\
&= \mR\,(\mK_t^{\top}\mP_{t-1}\mK_t + \mI_m)^{-1} \mK_t^{\top}\mP_{t-1},
\end{aligned}
\]
which proves the claim.
\end{proof}

\section{Detailed Complexity Derivations}\label{app:complexity}

\subsection{Preliminaries and Notation}
Let $d$ be the hidden size of the edited layer. 
At edit step $t$, denote the current key matrix $\mK_t\!\in\!\R^{d\times k_t}$, the stacked preserved/previous keys $\mK_p\!\in\!\R^{d\times m_p}$, and the residual $R_t$. 
We implement the projector in operator form $\mP = I - \mQ\mQ^\top$ with $\mQ\!\in\!\R^{d\times r}$ (rank $r\!\ll\!d$). 
All complexities below are per edit; we count only the dominant FLOPs and omit lower-order terms.

\subsection{\texttt{AlphaEdit} — Full Derivation and Cost}\label{app:alphaedit}
\texttt{AlphaEdit} uses the projected closed form
\begin{equation}
\Delta_{\text{Alpha}} \;=\; 
\mR_t\,\mK_t^\top \mP\,
\left( \mK_p\mK_p^\top \mP \;+\; \mK_t\mK_t^\top \mP \;+\; I \right)^{-1}.
\end{equation}
Define the $d\times d$ bracket as 
$\mM \!:=\! \mK_p\mK_p^\top \mP + \mK_t\mK_t^\top \mP + I$. 
Assuming $\mP$ is applied as an operator $\left(I{-}\mQ\mQ^\top\right)$:
\begin{itemize}
\item \textbf{Forming $\mM$.} Two Gram terms plus the low-rank projector correction: 
$O\!\left(d^2(m_p{+}k_t)\right) + O\!\left(d^2{\cdot}r\right)$.
\item \textbf{Inverting $\mM$.} Cholesky/LU on $d\times d$: $O\!\left(d^3\right)$.
\item \textbf{Assembly.} $\mR_t\mK_t^\top \mP\,\mM^{-1}$ is non-dominant after inversion ($O\!\left(d^2{\cdot}k_t\right)$).
\end{itemize}
\noindent\textbf{Per-edit cost.}
\begin{equation}
\boxed{~O\!\left(d^3\right)\;+\;O\!\left(d^2\left(m_p{+}k_t\right)\right)~},
\end{equation}
i.e., the cubic term is tied to $d$, while the pre-inversion work grows linearly with the accumulated size $m_p$.

\paragraph{Remark (no Woodbury).}
\texttt{AlphaEdit} inverts a $d\times d$ system directly. 
Replacing $\mP$ by a dense $d\times d$ projector would increase the forming cost to $\Theta\left(d^3\right)$ as well; hence operator form is strongly preferable even in \texttt{AlphaEdit}, though it does not change the cubic dependence on $d$.

\subsection{\methodname{} (Alg.\ 2: Steps 5, 7, and 12) — Full Derivation and Cost}\label{app:evoedit}
\methodname{} avoids any $d\times d$ inversion via projector alignment and a small inner system.

\paragraph{Step 5 (Alignment SVD).}
Compute $\mZ \!=\! \mP_{t-2}\mK_{t-1}$ in operator form and take a thin SVD of $\mZ\!\in\!\R^{d\times k_{t-1}}$ to extract dominant left singular vectors $\mQ_{t-1}$:
\[
\mZ \leftarrow \mK_{t-1} - \mQ\left(\mQ^\top \mK_{t-1}\right) \quad\Rightarrow\quad O\!\left(d\,r\,k_{t-1}\right),
\]
\[
\mZ = \mU\bm{\Sigma} \mV^\top,\quad \mQ_{t-1} \leftarrow \mU[:,1{:}r] \quad\Rightarrow\quad O\!\left(d\,k_{t-1}^2\right).
\]

\paragraph{Step 7 (Projector Update).}
Update $\mP_{t-1}$ in operator form by deflation
\[
\mP_{t-1} \;=\; \mP_{t-2} - \mQ_{t-1}\mQ_{t-1}^\top.
\]
In operator storage, this is a metadata refresh ($O\!\left(1\right)$). 
(Materializing a dense $\mP$ would cost $O\!\left(d^2r\right)$ per update and is discouraged.)

\paragraph{Step 12 (Small Inner System via Woodbury-style Rearrangement).}
Starting from the closed form
\[
\bm{\Delta}_{t}\mP_{t-1} \;=\; \mR_t \mK_t^\top \mP_{t-1}\,\left(\mK_t\mK_t^\top \mP_{t-1} + I\right)^{-1},
\]
apply the identity $\left(I_d{+}\mA\mB\right)^{-1}\mA = \mA\left(I_m{+}\mB\mA\right)^{-1}$ with $\mA{=}\mK_t$ and $\mB{=}\mK_t^\top \mP_{t-1}$, which yields a $k_t\times k_t$ inner system.
The resulting compute is:
\[
\mY = \mP_{t-1}\mK_t \Rightarrow O\!\left(d\,r\,k_t\right),\quad 
\mM = \mK_t^\top \mY \Rightarrow O\!\left(d\,k_t^2\right),
\]
\[
\text{factorize/solve } \left(I_{k_t}+\mM\right) \Rightarrow O\!\left(k_t^3\right),
\quad \bm{\Delta}_{t}\mP_{t-1} = \mR_t \mX \;(\text{assembly}).
\]

\paragraph{Per-edit cost (\methodname{}).}
\begin{equation}
\boxed{~O\!\left(d{\cdot}r{\cdot}k_{t-1}\right) + O\!\left(d{\cdot}k_{t-1}^2\right) + O\!\left(d{\cdot}r{\cdot}k_t\right) + O\!\left(d{\cdot}k_t^2\right) + O\!\left(k_t^3\right)~}.
\end{equation}
The only cubic term is $O\!\left(k_t^3\right)$, which depends on the current edit size $k_t$ (typically $k_t \ll d$) and is independent of the accumulated $m_p$.

\paragraph{Implementation Notes.}
\begin{enumerate}[leftmargin=1.5em,itemsep=0.1em,topsep=0.2em]
\item \textbf{Operator projector.} Always store $\mP$ as $I{-}\mQ\mQ^\top$, never as a dense $d\times d$ matrix. 
Every application $\mP\mX$ becomes $\mX - \mQ\left(\mQ^\top \mX\right)$ with cost $O\!\left(d{\cdot}r{\cdot}\mathrm{cols}\left(\mX\right)\right)$ instead of $O\!\left(d^2{\cdot}\mathrm{cols}\left(\mX\right)\right)$.
\item \textbf{Thin/Truncated SVD.} Since $k_{t-1}\!\ll\!d$, the SVD in Step~5 is cheap and numerically stable. Truncation (threshold $\tau$) controls $r$ and the downstream projector cost.
\item \textbf{Inner solve.} Use Cholesky on $\left(I_{k_t}{+}\mM\right)$ for stability; forward/backward substitutions dominate the assembly FLOPs and are already counted in $O\!\left(d{\cdot}k_t^2\right)$.
\end{enumerate}
\subsection{Side-by-side Takeaway}
\texttt{AlphaEdit} ties the cubic work to $d$ and grows linearly with $m_p$ in the forming stage, 
whereas \methodname{} binds the cubic term to $k_t$ and decouples from $m_p$ by design (projector alignment $+$ inner system). 
In the practical regime $k_t\ll d$ and $r\ll d$, \methodname{} offers a substantial and persistent per-edit advantage.
\begin{table}[ht!]
\centering\small
\caption{{Per-edit computational complexity.} \methodname{} reduces the cubic dependency from the hidden size $d_K$ to the much smaller edit size $n_t$, yielding significantly lower computational cost.}
\label{tab:complexity}
\resizebox{0.6\columnwidth}{!}{%
\begin{tabular}{l c c}
\toprule
Method & Dominant per-edit cost & Cubic term\\
\midrule
\makecell[l]{\texttt{AlphaEdit}} 
& \makecell[c]{\rule{0pt}{2.2ex}$O\!\left(d_K^3\right)+O\!\left(d_K^2(N{+}n_t)\right)$}
& \makecell[c]{$d_K$} \\
\midrule
\makecell[l]{\methodname{} (ours)}
& \makecell[c]{\rule{0pt}{2.2ex}$O\!\left(d_K{\cdot}r{\cdot}n_{t-1}\right) + O\!\left(d_K{\cdot}n_{t-1}^2\right)$\\
$O\!\left(d_K{\cdot}r{\cdot}n_t\right) + O\!\left(d_K{\cdot}n_t^2\right) + O\!\left(n_t^3\right)$}
& \makecell[c]{$n_t$} \\
\bottomrule
\end{tabular}%
}
\end{table}
\subsection{Detailed algorithms for \methodname}

\begin{algorithm}[th!]
\caption{Efficient Computation of $\bm{\Delta}_t \mP_{t-1}$ via the Woodbury Identity}
\label{alg:delta_computation}
\begin{algorithmic}[1]
\Require Projection matrix $\mP_{t-1}\!\in\!\R^{d \times d}$,\\
Current key matrix $\mK_t\!\in\!\R^{d \times r}$,\\
Residual matrix $\mR_t\!\in\!\R^{m \times d}$.
\Ensure
Updated term $\bm{\Delta}_t \mP_{t-1}$.
\vspace{3pt}
\State $\mY \gets \mP_{t-1} \mK_t$ \Comment{Project $\mK_t$ through $\mP_{t-1}$}
\State $\mM \gets \mK_t^\top \mY$ \Comment{Form the small $r \!\times\! r$ matrix}
\State $\mS \gets \mI_r + \mM$ \Comment{Compute the inner system}
\State Factor $\mS = \mL \mL^\top$ \Comment{Stable Cholesky Decomp.}
\State Solve $\mL \mZ = \mY^\top$ for $\mZ$ \Comment{Forward Sub.}
\State Solve $\mL^\top \mX = \mZ$ for $\mX$ \Comment{Backward Sub.}
\State $\bm{\Delta}_t \mP_{t-1} \gets \mR_t \mX$ \Comment{Final update}

\end{algorithmic}
\end{algorithm}

\begin{algorithm}[th!]
\caption{\methodname: Sequential Editing via Evolving Null-space Alignment}
\label{alg:evoedit}
\begin{algorithmic}[1]
\Require Initial weights $\mW_0$ and projector $\mP_0$, Data sequence $\{(s_1, r_1, o_1), \dots, (s_T, r_T, o_T)\}$
\For{$t = 1$ to $T$}
    \State Extract $(\mK_t, \mV_t)$ with $(s_t, r_t, o_t)$. 
    \If{$t > 1$} 
        \State Project $\mZ \gets \mP_{t-2}\mK_{t-1}$
        \State Decompose $\left(\mU_{t-1},\!\bm{\Sigma},\!\mV_{t-1}\right)\!\gets\!\mathrm{SVD}\!\left(\mZ\right)$
        \State Extract singular vectors with large singular values $\mQ_{t-1} = {\mU_{t-1}}_{[:,: d]}$
        \State Update $\mP_{t-1} \gets \mP_{t-2} - \mQ_{t-1}\mQ_{t-1}^\top$
    \Else 
        \State $\mP_{t-1} \gets \mP_{0}$
    \EndIf
    \State Compute residual: $\mR_t \gets \mV_t\!-\!\mW_{t-1}\mK_t$
    \State Solve closed-form update (\cref{eq:solution2}):
    \[
    \bm{\Delta}_t\!\mP_{t-1} = \mR_t\left(\mI + \mK_t^\top \mP_{t-1}\mK_t\right)^{-1}\mK_t^\top \mP_{t-1}
    \]
    \State Update weights: $\mW_t \gets \mW_{t-1} + \bm{\Delta}_t\!\mP_{t-1}$
\EndFor
\Ensure Updated model $\mW_T$
\end{algorithmic}
\end{algorithm}
\newpage
\twocolumn
\section{Experimental Details}\label{app:experiment_details}

\subsection{Datasets}
Here we provide a detailed introduction of the datasets used in our experiments:
\begin{itemize}
    \item \textbf{CounterFact}\cite{ROME}: A widely used benchmark for knowledge editing. Compared to other datasets, it is more challenging and explicitly contrasts counterfactual with factual statements. The benchmark evaluates efficacy, generalization, specificity, consistency, and fluency, and includes records spanning diverse subjects, relations, and linguistic forms
   \item \textbf{ZsRE}~\cite{Zsre}: A QA dataset whose questions are augmented via back-translation to create paraphrastic neighbors. Following prior work, Natural Questions is used as out-of-scope data to assess locality. Each example includes a subject string and answer for evaluating success of edits, a rephrased question for generalization, and a locality question for specificity.
\end{itemize}
\subsection{Metrics}
We evaluate on \textbf{CounterFact} with five metrics:

\begin{itemize}
  \item \textbf{Efficacy (efficacy success).}
  The proportion of cases in which the edited object achieves higher probability than the counterfactual on the original prompt:
  \[
  \mathbb{E}_i\!\big[\, \mathbb{P}_{f_\theta}[o_i \mid (s_i,r_i)] \;>\; \mathbb{P}_{f_\theta}[o_c^{i} \mid (s_i,r_i)] \,\big].
  \]

  \item \textbf{Generalization (paraphrase success).}
  The success rate computed on paraphrased prompts, averaging the per-item indicator over the paraphrase set:
  \[
  \mathbb{E}_i\!\big[\, \mathbb{P}_{f_\theta}[o_i \mid N((s_i,r_i))] \;>\; \mathbb{P}_{f_\theta}[o_c^{i} \mid N((s_i,r_i))] \,\big],
  \]
  where $N(\cdot)$ denotes paraphrased statements.

  \item \textbf{Specificity (neighborhood success).}
  The proportion of neighborhood prompts on which the model achieves higher probability to the edited fact:
  \[
  \mathbb{E}_i\!\big[\, \mathbb{P}_{f_\theta}[o_i \mid O\!((s_i,r_i))] \;>\; \mathbb{P}_{f_\theta}[o_c^{i} \mid O\!((s_i,r_i))] \,\big],
  \]
  where $O\!(\cdot)$ is the neighbor set.

   \item \textbf{Fluency (generation entropy).}
  Measure of repetition in model outputs, computed with a weighted entropy of bigrams and trigrams and averaged across outputs.
  \[
  -\frac{2}{3}\sum_{k} g_{2}(k)\log_{2} g_{2}(k)\;+\;\frac{4}{3}\sum_{k} g_{3}(k)\log_{2} g_{3}(k),
  \]
  where $g_n(\cdot)$ is the $n$\nobreakdash-gram frequency distribution.

  \item \textbf{Consistency (reference score).}
  Cosine similarity between TF–IDF vectors of the model text for subject $s$ and a reference description of $o$, averaged over items.
\end{itemize}
\begin{table*}[t!]
    \centering
    \resizebox{1.\linewidth}{!}{\begin{tabular}{cc|ccccc|ccc}
\toprule[1.5pt]
\multirow{2}{*}{\textbf{Method}} & \multirow{2}{*}{{\textbf{Model}}}  & \multicolumn{5}{c|}{\textsc{\textbf{Counterfact}}} & \multicolumn{3}{c}{\textsc{\textbf{ZsRE}}} \\
\cmidrule(lr){3-7} \cmidrule(lr){8-10}
&& \textbf{Eff.$\uparrow$} & \textbf{Gen.$\uparrow$} & \textbf{Spe.$\uparrow$} & \textbf{Flu.$\uparrow$} & \textbf{Consis.$\uparrow$} & \textbf{Eff.$\uparrow$} & \textbf{Gen.$\uparrow$} & \textbf{Spe.$\uparrow$} \\
\midrule[1pt]

\texttt{ROME}  & \multirow{4}{*}{\rotatebox{90}{\texttt{GPT2-XL}}} & {54.60\std{0.48}} & {51.18\std{0.40}} & {52.68\std{0.33}} & {366.13\std{1.40}} & {0.72\std{0.02}} & {47.50\std{0.43}} & {43.56\std{0.42}} & {14.27\std{0.19}}\\
\texttt{MEMIT} & & {94.70\std{0.22}} & {85.82\std{0.28}} & {60.50\std{0.32}} & {477.26\std{0.54}} & {22.72\std{0.15}} & {79.17\std{0.32}} & {71.44\std{0.36}} & {26.42\std{0.25}}\\
\texttt{AlphaEdit} & & \textbf{99.50\std{0.24}} & \textbf{93.95\std{0.34}} & \underline{66.39\std{0.31}} & \textbf{597.88\std{0.18}} & \textbf{39.38\std{0.15}} & \textbf{94.81\std{0.30}} & \underline{86.11\std{0.29}} & \underline{25.88\std{0.21}}\\
\methodname{} (Ours) & & \underline{99.32\std{0.16}} & \underline{92.96\std{0.43}} & \textbf{66.53\std{0.12}} & \underline{592.04\std{1.12}} & \underline{38.04\std{0.22}} & \underline{94.51\std{1.93}} & \textbf{87.70\std{2.47}} & \textbf{25.89\std{0.07}}\\
\midrule[1pt]

\texttt{ROME}   & \multirow{4}{*}{\rotatebox{90}{\texttt{GPT-J-6B}}} & {57.50\std{0.48}} & {54.20\std{0.40}} & {52.05\std{0.31}} & {589.42\std{0.08}} & {3.22\std{0.02}} & {56.42\std{0.42}} & {54.65\std{0.42}} & {9.86\std{0.16}}\\
\texttt{MEMIT}  & & {98.55\std{0.11}} & {{95.50\std{0.16}}} & {63.64\std{0.31}} & {546.28\std{0.88}} & {34.89\std{0.15}} & {94.91\std{0.16}} & {90.22\std{0.23}} & {\textbf{30.39\std{0.27}}}\\
\texttt{AlphaEdit} & & \underline{99.75\std{0.08}} & \textbf{96.38\std{0.23}} & \textbf{75.48\std{0.21}} & \underline{618.50\std{0.17}} & \textbf{42.08\std{0.15}} & \textbf{99.79\std{0.14}} & \underline{96.00\std{0.22}} & {28.29\std{0.25}}\\
\methodname{} (Ours) & & \underline{99.75\std{0.00}} &\underline{95.94\std{0.28}} & \underline{75.21\std{0.53}} & \textbf{619.12\std{1.23}} & \underline{41.42\std{0.27}} & \underline{98.69\std{0.21}} & \textbf{96.46\std{0.27}} & \underline{28.62\std{0.21}}\\
\bottomrule
    \end{tabular}}
    \caption{Evaluation results of GPT2-XL and GPT-J-6B for \methodname{} and existing methods.}
    \label{tab:placeholder}
\end{table*}
For the \textbf{ZsRE} dataset, we evaluate on three metrics:

\begin{itemize}
  \item \textbf{Efficacy.}
  Average top-1 accuracy on the edit prompts:
  \[
  \mathbb{E}_i\!\Big\{\, o_i \;=\; \arg\max_{o}\; \mathbb{P}_{f_\theta}\!\big(o \mid (s_i,r_i)\big) \Big\}.
  \]

  \item \textbf{Generalization.}
  Performance on paraphrased variants of the same fact, denoted by $N((s_i,r_i))$:
  \[
  \mathbb{E}_i\!\Big\{\, o_i \;=\; \arg\max_{o}\; \mathbb{P}_{f_\theta}\!\big(o \mid N((s_i,r_i))\big) \Big\}.
  \]

  \item \textbf{Specificity.}
  Ensures the edit does not affect unrelated prompts $O\!((s_i,r_i))$; measured by the top-1 accuracy of \emph{unchanged} predictions:
  \[
  \mathbb{E}_i\!\Big\{\, o_i^{c} \;=\; \arg\max_{o}\; \mathbb{P}_{f_\theta}\!\big(o \mid O\!((s_i,r_i))\big) \Big\}.
  \]
\end{itemize}
\subsection{Baselines}
We compare \methodname{} against three representative editing methods:

\begin{itemize}
  \item \texttt{ROME}~\cite{ROME}. A method for updating specific factual associations in LLMs. It identifies key activations in mid-layer feed-forward modules that mediate factual predictions and adjusts their weights to favor the edited object while preserving locality.
  \item \texttt{MEMIT}~\cite{MEMIT}. A scalable multi-layer update algorithm for inserting many new memories into transformer models. It builds on \texttt{ROME}, coordinates low-rank edits across several layers, and uses an efficient closed-form solver suitable for large batches of facts.
  \item \texttt{AlphaEdit}~\cite{Alphaedit}. A sequential editor for long runs of updates. For each new fact it projects the change into a subspace that preserves prior knowledge in the model.
\end{itemize}
\section{Additional Experimental Results on GPTJ and GPT2-XL}
\section{Additional Experimental Results with LangEdit}
\begin{table}[th!]
\centering
\resizebox{1.\linewidth}{!}{
\begin{tabular}{lccccc}
\hline
\textbf{llama3-8B} & \textbf{Eff.} & \textbf{Gen.} & \textbf{Spe.} & \textbf{Flu.} & \textbf{Consis.} \\
\hline
LangEdit & 50.25 & 49.98 & 49.97 & 521.14 & 0.85 \\
\methodname{}  & 99.67 & 94.93 & 69.99 & 623.09 & 32.64 \\
\hline
\end{tabular}}
\caption{Model editing performance on llama3-8B with LangEdit.}
\label{tab:llama3_edit}
\end{table}
Regarding \textit{LangEdit}, we intentionally omitted it from the main experiments because, in our setting, it collapses quickly as the number of edits increases. To make this explicit, we evaluated LangEdit on 2,000 sequential edits using llama3-8B on CounterFact. As shown below, LangEdit fails to maintain fidelity across all metrics, confirming that it is not stable in the large-scale sequential editing regime considered in our work.

\subsection{Detailed Hyperparameter Settings}
For \texttt{GPT2-XL}, \texttt{GPT-J-6B}, and \texttt{Llama-3-8B}, we follow the hyperparameters reported by \texttt{AlphaEdit}.
In \methodname{}, we additionally introduce a regularization coefficient $L_2$ multiplying the identity in \cref{eq:solution2}; we report its value alongside the base settings.

\begin{itemize}
  \item \texttt{GPT2-XL}. Edit layers $[13,14,15,16,17]$; $\lambda=20{,}000$; $20$ optimization steps; learning rate $0.5$; $L_2{=}1$.
  \item \texttt{GPT-J-6B}. Edit layers $[3,4,5,6,7,8]$; $\lambda=15{,}000$; $25$ optimization steps; learning rate $0.5$; $L_2{=}1$.
  \item \texttt{Llama-3-8B}. Edit layers $[4,5,6,7,8]$; $\lambda=15{,}000$; $25$ optimization steps; learning rate $0.1$; $L_2{=}4$.
  \item \texttt{Llama-3.2-3B}. Edit layers $[2,3,4,5,6]$; $\lambda=15{,}000$; $25$ optimization steps; learning rate $0.1$; $L_2{=}3$.
  \item \textbf{Qwen2.5-7B-Instruct}. Edit layers $[7,8,9,10,11]$; $\lambda=15{,}000$; $25$ optimization steps; learning rate $0.1$; $L_2{=}1$.
\end{itemize}

\end{document}